\documentclass{article}

\usepackage[T1]{fontenc}
\usepackage{lmodern}
\usepackage{amsmath}
\usepackage{amssymb}
\usepackage{graphicx}
\usepackage{array}
\usepackage{booktabs}
\usepackage{placeins}
\usepackage[font=small,labelfont=bf]{caption}
\usepackage[round,semicolon]{natbib}
\usepackage{xcolor}
\usepackage{hyperref}
\usepackage{doi}  %
\definecolor{linkblue}{rgb}{0 0.4235 0.6392}
\hypersetup{colorlinks, linkcolor=linkblue, citecolor=linkblue, urlcolor=linkblue,
  pdftitle={English Calibration Erodes Lithuanian in Expert-Pruned Mixture-of-Experts Language Models},
  pdfauthor={Tomas Rasymas, Povilas Kvedaras}}

\makeatletter
\newcommand{\V}[1]{\ifcsname V@#1\endcsname\csname V@#1\endcsname\else
  \textbf{\textcolor{red}{??#1??}}\PackageWarning{numbers}{Undefined value #1}\fi}
\makeatother
\providecommand{\defV}[2]{\expandafter\def\csname V@#1\endcsname{#2}}
\defV{ppl-qthree-lt}{5.7}
\defV{ppl-qthree-en}{7.4}
\defV{params-25-min}{22.1}
\defV{params-25-max}{23.7}
\defV{params-50-min}{44.3}
\defV{params-50-max}{47.5}
\defV{prunetime-min-min}{15}
\defV{prunetime-min-max}{52}
\defV{never-n}{30}
\defV{never-max}{17.1}
\defV{never-en-25}{4}
\defV{never-lt-25}{7}
\defV{never-qthree-n}{9}
\defV{never-qthree-max}{17.1}
\defV{never-qsix-n}{8}
\defV{never-qsix-max}{0.9}
\defV{never-gemma-n}{12}
\defV{never-gemma-max}{8.5}
\defV{never-glm-n}{1}
\defV{never-glm-max}{0.1}
\defV{sigpos-bel-en}{14}
\defV{sigpos-bel-lt}{11}
\defV{sigpos-bel-mix}{10}
\defV{sigpos-flo-en}{12}
\defV{sigpos-flo-lt}{2}
\defV{signeg-flo-lt}{6}
\defV{sigpos-flo-mix}{3}
\defV{signeg-flo-mix}{3}
\defV{sigpos-wiki-en}{16}
\defV{signeg-wiki-lt}{16}
\defV{signeg-wiki-mix}{11}
\defV{resid-glm-mix}{9.1}
\defV{resid-qthree-lt}{4.3}
\defV{resid-qsix-lt}{1.9}
\defV{resid-qsix-mix}{3.1}
\defV{encal25-med-en}{1.7}
\defV{encal25-med-lt}{21.4}
\defV{encal25-lt-max}{67.6}
\defV{q3euen25-en}{93.3}
\defV{q3euen25-en-loss}{1.0}
\defV{q3euen25-lt}{22.3}
\defV{q3euen25-lt-base}{89.9}
\defV{q3wien25-lt}{67.4}
\defV{glmeuen25-lt}{40.4}
\defV{glmwien25-lt}{54.2}
\defV{gwien25-en-loss}{19.7}
\defV{gwien25-gap}{+3.2}
\defV{g50-lt-min}{22}
\defV{g50-lt-max}{25}
\defV{g50-en-chance-n}{4}
\defV{eff-lten-bel-lt-better}{15}
\defV{eff-lten-bel-lt-worse}{0}
\defV{eff-lten-enlt-better}{16}
\defV{eff-lten-lten-better}{16}
\defV{eff-lten-wiki-en-better}{5}
\defV{eff-lten-wiki-en-worse}{11}
\defV{eff-lten-wiki-lt-better}{16}
\defV{eff-mixen-bel-en-worse}{1}
\defV{eff-mixlt-bel-lt}{\ensuremath{-}3.0}
\defV{eff-mixlt-bel-lt-worse}{7}
\defV{eff-mixlt-enlt}{\ensuremath{-}1.0}
\defV{eff-mixlt-enlt-worse}{8}
\defV{eff-mixlt-wiki-en-better}{12}
\defV{eff-mixlt-wiki-lt-worse}{14}
\defV{relnll-en-25-en}{5.2}
\defV{relnll-en-25-lt}{55.2}
\defV{relnll-en-50-en}{17.6}
\defV{relnll-en-50-lt}{134.3}
\defV{relnll-lt-25-en}{5.6}
\defV{relnll-lt-25-lt}{3.2}
\defV{relnll-lt-50-en}{22.1}
\defV{relnll-lt-50-lt}{14.0}
\defV{ltcal25-flo-maxdrop}{1.1}
\defV{q6lt50-enlt-min}{53.5}
\defV{q6lt50-enlt-max}{53.6}
\defV{q6-enlt-base}{54.0}
\defV{ltcal50-nogemma-lt-loss-min}{6}
\defV{ltcal50-nogemma-lt-loss-max}{15}
\defV{best-lt-bel-en}{0.9}
\defV{best-lt-bel-lt}{0.8}
\defV{best-mix-bel-en}{\ensuremath{-}0.3}
\defV{best-mix-bel-lt}{0.9}
\defV{ratio-bel-en-worse}{22}
\defV{ratio-bel-en-n}{24}
\defV{ratio-bel-lt-worse}{21}
\defV{ratio-bel-lt-n}{24}
\defV{ratio-enlt-worse}{24}
\defV{ratio-enlt-n}{24}
\defV{ratio-lten-worse}{24}
\defV{ratio-lten-n}{24}
\defV{ratio-wiki-en-worse}{24}
\defV{ratio-wiki-en-n}{24}
\defV{ratio-wiki-lt-worse}{24}
\defV{ratio-wiki-lt-n}{24}
\defV{ratio-loss-x-min}{2.4}
\defV{ratio-loss-x-max}{6.9}
\defV{ds-bel-en-europarl}{12}
\defV{ds-bel-en-wikipedia}{1}
\defV{ds-lten-europarl}{3}
\defV{ds-lten-wikipedia}{12}
\defV{ds-wiki-en-wikipedia}{24}
\defV{ds-wiki-en-n}{24}
\defV{ds-wiki-lt-wikipedia}{20}
\defV{ds-wiki-lt-n}{24}
\defV{sal-enlt-min}{0.72}
\defV{sal-enlt-max}{0.85}
\defV{selent-lt-min}{0.77}
\defV{selent-lt-max}{0.82}
\defV{selent-en-min}{0.86}
\defV{selent-en-max}{0.92}
\defV{jsd-min}{0.15}
\defV{jsd-max}{0.26}
\defV{jac-enlt-25-min}{0.81}
\defV{jac-enlt-25-max}{0.88}
\defV{shared-enlt-25-min}{90}
\defV{shared-enlt-25-max}{94}
\defV{jac-random-25}{0.60}
\defV{jac-enlt-50-min}{0.64}
\defV{jac-enlt-50-max}{0.78}
\defV{jac-random-50}{0.33}
\defV{remjac-enlt-25-min}{0.52}
\defV{remjac-enlt-25-max}{0.68}
\defV{remjac-random-25}{0.14}
\defV{remmass-en-25-en}{13.6}
\defV{remmass-en-25-lt}{20.2}
\defV{remmass-en-50-en}{37.2}
\defV{remmass-en-50-lt}{46.7}
\defV{remmass-lt-25-en}{15.3}
\defV{remmass-lt-25-lt}{12.2}
\defV{remmass-ratio}{1.5}
\defV{depth-first-jsd-min}{0.16}
\defV{depth-first-jsd-max}{0.28}
\defV{depth-middle-jsd-min}{0.09}
\defV{depth-middle-jsd-max}{0.13}
\defV{depth-middle-rho-min}{0.84}
\defV{depth-middle-rho-max}{0.91}
\defV{depth-last-jsd-min}{0.33}
\defV{depth-last-jsd-max}{0.49}
\defV{depth-last-top-en-min}{19}
\defV{depth-last-top-en-max}{39}
\defV{depth-last-top-lt-min}{55}
\defV{depth-last-top-lt-max}{72}
\defV{bandmass-en-last-en}{10.3}
\defV{bandmass-en-last-lt}{20.5}
\defV{extra-qsix-last}{+12.9}
\defV{extra-qsix-middle}{\ensuremath{-}0.3}
\defV{extra-gemma-late}{+19.7}
\defV{extra-glm-last}{+13.8}
\defV{extra-glm-middle}{+5.9}
\defV{extra-qthree-min}{3}
\defV{extra-qthree-max}{6}
\defV{ovl-bel-pooled-rho}{0.06}
\defV{ovl-bleu-pooled-rho}{0.62}
\defV{ovl-wiki-pooled-rho}{\ensuremath{-}0.59}
\defV{ovl-25-maxabs}{0.11}
\defV{hits-glm-en25-min}{27}
\defV{hits-glm-en25-max}{36}
\defV{hits-glm-euen50}{94}
\defV{g50-A-min}{46}
\defV{g50-A-max}{95}
\defV{g50-A-n}{12}
\defV{g50-mass-min}{0.01}
\defV{g50-mass-max}{0.25}
\defV{encal50-lt-A-min}{78}
\defV{encal50-lt-A-max}{99}
\defV{collapsed-top-min}{72}
\defV{healthy-top-max}{41}
\defV{thought-50-configs}{6}
\defV{thought-50-lten-min}{10}
\defV{thought-50-lten-max}{712}
\defV{a1-sal-min}{0.77}
\defV{a1-sal-max}{0.93}
\defV{a1-jsdextra-min}{0.25}
\defV{a1-jsdextra-max}{0.44}
\defV{a2-rho}{0.95}
\defV{a2-mcnemar-raw}{83}
\defV{a2-mcnemar-bh}{81}
\defV{a2-n}{96}
\defV{a3-rho-rel}{0.91}
\defV{a3-dis-abs}{12}
\defV{a4-en-25-enlt-hits}{20.2}
\defV{a4-en-25-enlt-all}{22.7}
\defV{a4-en-25-enlt-kept}{16.7}
\defV{a4-en-25-enlt-recovered}{2.3}
\defV{a4-g-europarllt50-lten-all}{12.5}
\defV{a4-g-europarllt50-lten-nothought}{10.2}
\defV{a9-ltcal25-enlt-chrfdrop}{1.1}
\defV{a5-ltmix-chosen}{8}
\defV{a5-n}{8}
\defV{a5-encal25-en-within2}{5}
\defV{a6-all50-qthree}{16.7}
\defV{a6-all50-qsix}{11.5}
\defV{a6-all50-gemma}{66.4}
\defV{a6-ltgini-higher-pct}{61}
\defV{a6-ltgini-higher-qthree}{83}
\defV{a6-ltgini-higher-qsix}{36}
\defV{a6-ltgini-higher-gemma}{68}
\defV{a6-ltgini-higher-glm}{55}
\defV{a7-sel-min}{0.016}
\defV{a7-sel-max}{0.041}
\defV{off-base-max}{0.0}
\defV{off-en-25-enlt}{3.9}
\defV{off-en-50-enlt}{59.4}
\defV{off-en-50-enlt-max}{95.0}
\defV{off-lt-50-enlt}{10.9}
\defV{off-mix-50-enlt}{14.1}
\defV{split-en-25-eng}{3.4}
\defV{split-en-25-ltlike}{32.6}
\defV{split-en-50-eng}{55.3}
\defV{split-lt-25-ltlike}{6.2}
\defV{top8-last-encal-lt-min}{0.86}
\defV{top8-last-encal-lt-max}{2.88}
\defV{top8-last-ltcal-lt-min}{0.00}
\defV{top8-last-ltcal-lt-max}{0.71}
\defV{cov-slope-en}{\ensuremath{-}0.12}
\defV{h3-belebele-en-up}{3}
\defV{h3-belebele-en-down}{2}
\defV{h3-belebele-lt-up}{3}
\defV{h3-flores-en-up}{3}
\defV{h3-flores-en-down}{4}
\defV{h3-wikipedia-en-up}{8}
\defV{h3-wikipedia-lt-down}{7}
\defV{strict-base-en}{91.5}
\defV{strict-base-lt}{83.4}
\defV{strict-en-25-en}{87.4}
\defV{acc-en-25-en}{88.6}
\defV{strict-en-25-lt}{44.2}
\defV{acc-en-25-lt}{58.9}
\defV{binom-p}{0.065}
\defV{item-lt-25-en}{3.8}
\defV{item-lt-25-lt}{8.5}
\defV{item-lt-25-ltmore}{8}
\defV{eta-ltdeg-model}{22}
\defV{eta-ltdeg-lang}{24}
\defV{eta-ltdeg-ratio}{21}
\defV{eta-belgap-lang}{39}
\defV{eta-flogap-lang}{36}
\defV{eta-wikigap-lang}{70}
\defV{ml-bel-lt-qthree}{49.8}
\defV{ml-bel-lt-qsix}{17.9}
\defV{ml-bel-lt-gemma}{5.1}
\defV{ml-bel-lt-glm}{29.3}
\defV{ml-enlt-gemma}{24.3}
\defV{ml-enlt-glm}{33.8}
\defV{content-ratio-min}{1.7}
\defV{content-ratio-max}{2.1}
\defV{limit-words-lt-min}{79}
\defV{limit-words-lt-max}{100}
\defV{limit-words-en-min}{201}
\defV{limit-words-en-max}{205}
\defV{seq-25-min}{0.85}
\defV{seq-25-max}{0.97}
\defV{seq-50-min}{0.82}
\defV{seq-50-max}{0.96}
\defV{cs-median}{0.36}
\defV{prefix-sized-pct}{48}
\defV{tau-centred-25}{0.25}
\defV{tau-centred-25-p}{0.09}
\defV{r12-en_all-extra_lt-bel}{\ensuremath{-}0.29}
\defV{r12-all_25-lt_first-ltdeg}{0.87}
\defV{r12-all_25-lt_last-ltdeg}{0.75}
\defV{r12-all_25-lt-ltdeg}{0.72}
\defV{eff25-lten-bel-lt}{22.3}
\defV{remmass-ratio-50}{1.26}
\defV{encal25-within2-lt-min}{19.3}
\defV{encal25-within2-lt-max}{67.6}
\defV{g50-A-ge70}{11}
\defV{ds-wiki-signed}{0.15}
\defV{ds-bel-lt-lt-sig}{4}
\defV{ds-bel-lt-lt-maxabs}{7.1}
\defV{ds-bel-lt-mix-sig}{4}
\defV{ds-bel-lt-mix-maxabs}{22.0}
\defV{ni-worst-lo}{\ensuremath{-}4.9}
\defV{h5-worst-mix-better}{3}
\defV{h5-absgap-wiki-mix-smaller}{15}
\defV{gb-saved-25-min}{11.4}
\defV{gb-saved-25-max}{16.1}
\defV{wiki-median-bytes-en}{1{,}462}
\defV{wiki-median-bytes-lt}{924}
\defV{fffd-runs}{22}
\defV{fffd-total}{948}
\defV{fffd-glm-pct}{88}
\defV{sigpos-en-min}{12}
\defV{ratio-worse-min}{21}
\defV{gap-flo-en-25-u}{12.8}
\defV{gap-bel-lt-25-u}{3.0}
\defV{gap-bel-lt-50-u}{5.1}
\defV{gap-bel-mix-25-u}{5.6}
\defV{gap-bel-mix-50-u}{6.2}
\defV{eff-lten-bel-en-u}{2.5}
\defV{eff-lten-bel-lt-u}{25.5}
\defV{eff-lten-enlt-u}{28.2}
\defV{eff-lten-lten-u}{18.0}
\defV{eff-mixlt-bel-lt-u}{3.0}
\defV{eff-mixlt-enlt-u}{1.0}
\defV{ds-bel-en-mean-u}{3.1}
\defV{ds-lten-mean-u}{1.1}
\defV{cov-slope-lt-u}{1.85}
\defV{ltben-europarl-u}{29.7}
\defV{ltben-wikipedia-u}{21.3}
\defV{ni-median-u}{0.8}

\newcommand{\pp}{\,pp}
\newcommand{\figref}[1]{Figure~\ref{#1}}
\newcommand{\tabref}[1]{Table~\ref{#1}}
\newcommand{\appref}[1]{Appendix~\ref{#1}}
\newcolumntype{L}[1]{>{\raggedright\arraybackslash}p{#1}}
\title{English Calibration Erodes Lithuanian\\ in Expert-Pruned Mixture-of-Experts\\ Language Models}
\author{Tomas Rasymas\thanks{Corresponding author:
  \href{mailto:tomas.rasymas@hostinger.com}{tomas.rasymas@hostinger.com}.}\qquad Povilas Kvedaras\\[2pt]
  Hostinger Research}
\date{}

\begin{document}
\maketitle

\begin{abstract}
Expert pruning compresses Mixture-of-Experts (MoE) language models by removing the experts that a calibration set
marks as unimportant. Studies of high-resource languages largely find that English calibration transfers well to other
languages. We test this for Lithuanian: we pruned four MoE models with REAP, removing 25\% or 50\% of the experts with
English, Lithuanian or mixed calibration, and evaluated both languages on reading comprehension, translation and
language modelling. English calibration does not transfer to Lithuanian: at 25\% pruning, English comprehension fell
by a median of \V{encal25-med-en} points but Lithuanian by \V{encal25-med-lt}. The transfer is asymmetric. Lithuanian
calibration removed most of the Lithuanian loss and never significantly hurt English comprehension, which was on
average \V{eff-lten-bel-en-u} points better than under English calibration; only English language modelling got
slightly worse. Routing suggests why: in the last layers, Lithuanian sends
\V{depth-last-top-lt-min}--\V{depth-last-top-lt-max}\% of its expert selections to just eight experts, against
\V{depth-last-top-en-min}--\V{depth-last-top-en-max}\% for English, and English calibration removes some of these
experts. Compressed models should be calibrated, and evaluated, in every language they must serve.
\end{abstract}

\begin{quote}\small
\end{quote}

\section{Introduction}\label{sec:intro}

Mixture-of-Experts (MoE) models \citep{shazeer2017outrageously,fedus2022switch} send each token to a few of many
feed-forward ``experts''. The open MoE models we study have 26--35B parameters but use only 3--4B of them per token, so
they are cheap to run but expensive to store. \emph{Expert pruning} cuts this memory cost by deleting whole experts.
REAP (Router-weighted Expert Activation Pruning; \citealp{lasby2025reap}) scores each expert on a small calibration
set, by how much weight the router gives it and how large its output is, and removes the lowest-scoring experts; it
outperforms expert merging on generative tasks.

The calibration set is almost always English. The REAP authors showed that its \emph{domain} matters: calibrating on
general web text can collapse a model's code generation. Whether its \emph{language} matters is less clear. For MoE
pruning, studies of high-resource languages suggest that English calibration transfers well
\citep{salim2026expert,salim2026towards,martin2026extracting}. For dense models, English-calibrated compression is known
to hurt other languages more \citep{choenni2025mwanda,marie2025uneven,chimoto2026calibrating}.

We study Lithuanian, an East Baltic language and an official language of the European Union. Of the residents of
Lithuania, 84.6\% are ethnic Lithuanians, and 99.4\% of them name Lithuanian as their mother tongue
\citep{statisticslithuania2021census}. Lithuanian is richly inflected, with seven noun cases
\citep{jakubauskaite2026evaluating}, and its alphabet adds nine letters with diacritics (\k{a}, \v{c}, \k{e}, \.{e},
\k{i}, \v{s}, \k{u}, \={u}, \v{z}). It is also scarce on the web: 0.17\% of a recent Common Crawl archive, against
42.6\% for English \citep{commoncrawl2026languages}. The models we study need \V{content-ratio-min}--\V{content-ratio-max}
times as many tokens for a Lithuanian sentence as for its English translation \citep{petrov2023language}. Lithuanian
has recent open models \citep{nakvosas2025open,bergmanis2026tildeopen} and evaluations
\citep{kapociutedzikiene2025localizing,jakubauskaite2026evaluating}, but we found no study of how MoE compression
affects it.

\begin{figure}[ht]
\includegraphics[width=\textwidth]{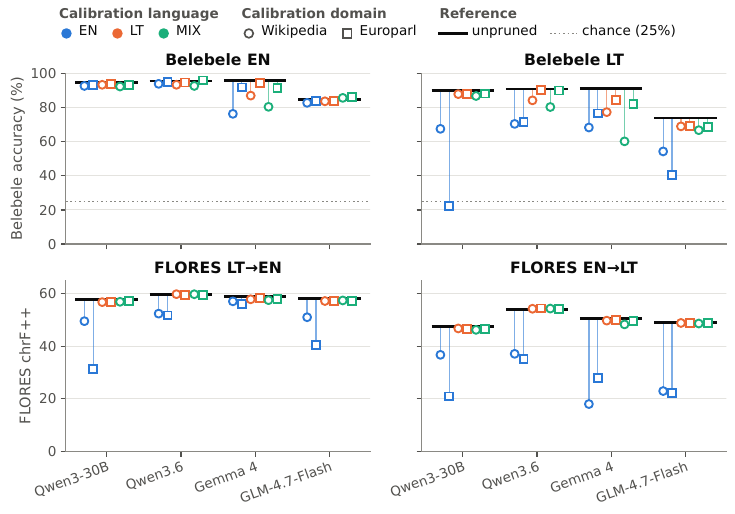}
\caption{\textbf{At 25\% expert pruning, English calibration erodes Lithuanian far more than English.} Black bars are
the unpruned models and points their 25\%-pruned versions. Unless a caption says otherwise, colour marks the
calibration language in all figures (EN blue, LT orange, MIX green); circles and squares mark Wikipedia and Europarl
calibration, and hollow and filled markers mark 25\% and 50\% pruning. Higher is better; the dotted line is chance on
Belebele. English-calibrated models keep their English comprehension (except Gemma~4 with English Wikipedia
calibration) but lose up to \V{encal25-lt-max} points in Lithuanian, and they translate worse, especially into
Lithuanian.}
\label{fig:headline}
\end{figure}

We prune four MoE models with REAP, removing 25\% or 50\% of the experts in every layer. The calibration text is
English (EN), Lithuanian (LT) or a 50/50 mix (MIX), from one of two domains: Wikipedia or the parallel Europarl corpus.
We then test both languages on reading comprehension (Belebele), translation in both directions (FLORES-200) and
language modelling (Wikipedia). We ask six research questions and, for the first three, test five hypotheses:
\begin{description}
\item[RQ1] Does REAP degrade Lithuanian more than English? \emph{H1: yes.}
\item[RQ2] Does the calibration language change this gap, and what does English lose under Lithuanian calibration?
\emph{H2: Lithuanian calibration reduces the Lithuanian loss. H3: Lithuanian or mixed calibration does so without
substantially hurting English. H4: a mix balances the two languages better than Lithuanian alone.}
\item[RQ3] How does the gap depend on the pruning ratio? \emph{H5: it grows from 25\% to 50\%.}
\item[RQ4] Does the calibration domain matter?
\item[RQ5] Are the experts that matter language-specific?
\item[RQ6] How do pruned models fail?
\end{description}
\tabref{tab:hyp} gives the decision rule for each hypothesis and its verdict. Our main findings are:
\begin{itemize}
\item \textbf{English calibration hurts Lithuanian far more than English.} At 25\% pruning, English-calibrated models
lose a median of \V{encal25-med-en} points of English reading-comprehension accuracy but \V{encal25-med-lt} points of
Lithuanian (\figref{fig:headline}), and their translation into Lithuanian suffers most. An English-only evaluation
would have missed this damage.
\item \textbf{Lithuanian or mixed calibration removes most of the gap} with little or no cost to English comprehension,
although English language modelling gets slightly worse. The calibration language matters more for the gap than any
other design factor, including the calibration domain.
\item \textbf{Pruning half of the experts is risky whatever the calibration.} Losses grow several-fold from 25\% to
50\%, and even Lithuanian-calibrated models then lose \V{ltcal50-nogemma-lt-loss-min}--\V{ltcal50-nogemma-lt-loss-max}
points in Lithuanian.
\item \textbf{The two languages share most experts, but not all.} English and Lithuanian calibration keep
\V{shared-enlt-25-min}--\V{shared-enlt-25-max}\% of the same experts, yet the experts removed under English
calibration had carried a larger share of Lithuanian than of English routing. The difference is largest in the last
layers, where Lithuanian relies on a few experts.
\item \textbf{Pruned models fail visibly in Lithuanian}: they do not stop, switch to English, invent word forms and
break diacritics.
\end{itemize}
Beyond these findings, the study provides a controlled grid of 48 pruned models with per-language paired tests,
including a content-matched comparison on parallel text, and practical advice for compressing MoE models that must
serve Lithuanian (Section~\ref{sec:discussion}).

\section{Related Work}\label{sec:related}

\paragraph{Calibration language in MoE expert pruning}
Three recent studies ask whether MoE expert pruning needs calibration data in the target language, and for
high-resource languages largely find that English transfers well. On gpt-oss-20b, the experts chosen with English and with
Traditional Chinese data are nearly the same and give the same MMLU accuracy \citep{salim2026expert}. Across seven
high-resource languages, English-only calibration is competitive at moderate sparsity but falls behind when pruning is
aggressive \citep{salim2026towards}. \citet{martin2026extracting} prune gpt-oss-20b and Qwen3-30B-A3B into translation
specialists; they find that languages share most experts, that high-resource languages need no in-language
calibration data, and that translation out of English is more fragile than translation into it. We see the same
overlap and the same direction, but for Lithuanian and generic English calibration the few experts that are not
shared matter a great deal (\tabref{tab:prior}).

\paragraph{MoE expert pruning and merging}
REAP \citep{lasby2025reap} showed that general (C4) calibration can collapse code generation, especially at 50\%
pruning; our study is the language counterpart of this result. EASY-EP \citep{dong2025domain} uses a similar score and
finds that a few in-domain examples activate a stable set of experts. \citet{liu2026score} describe REAP and related
scores as one family and show that routing frequency, and to a lesser degree gate weights, make expert scores depend on
the calibration data. Other methods reduce this dependence. AIMER \citep{liu2026aimer} ranks experts from their weights
alone. REAM \citep{jha2026ream} merges low-scoring experts into the experts REAP keeps and finds that the calibration mix
trades multiple-choice against generative accuracy. Mosaic Pruning \citep{hu2025mosaic} reports collapse on domains
missing from a single calibration corpus. MoEXBench \citep{benazir2026benchmarking} applies REAP to ten MoE models,
three of them ours; it finds that expert pruning causes most of the damage when compression methods are combined, and
that the choice of calibration data can more than halve the coding score of pruned Qwen3.6. \citet{jha2026half} prunes
two of our models, Qwen3.6 and Gemma~4, for coding and finds the damage outside the calibration domain, and
\citet{yamaguchi2026utility} prune four MoE models with biomedical text and find that general ability drops while
surface metrics barely move. Removing just three of the 6{,}144 experts can break Qwen3-30B-A3B \citep{su2025unveiling},
and language-specific expert pruning has a precedent in translation MoEs \citep{koishekenov2023memory}.

\paragraph{Calibration data in dense-model compression}
For dense models the evidence is clearer. Non-English and multilingual calibration lowers the perplexity of quantised
Llama and Qwen models \citep{chimoto2026calibrating}. Wanda pruning \citep{sun2024simple} at 50\% sparsity or more hurts
under-represented languages in ways that an English-only evaluation does not show \citep{choenni2025mwanda}, and pruned
models can pass multiple-choice items that they no longer answer in open generation: at 50\% Wanda sparsity, English
generation accuracy drops by 1.1 points but Swahili by 21.3 \citep{wen2026benchmark}. Layer-importance rankings depend
on the calibration language, and a mixed set is no better than English \citep{wibowo2025multilingual}. Quantisation
hurts low-resource translation most, and language-matched calibration helps at low bit-widths \citep{marie2025uneven}.
Calibration-language effects are large but differ between models \citep{he2025preserving}, and at high sparsity a
balanced mix of calibration sources beats any single source \citep{xu2026rethinking}. For MoE quantisation, MoEQuant
\citep{chen2025moequant} traces errors to rarely routed experts that see few calibration tokens. Earlier work
established that compression is sensitive to calibration data in general \citep{williams2024impact} and in
multilingual models \citep{zeng2024multilingual,kurz2026limitations,marchisio2024quantization}.

\paragraph{Multilingual routing and language-specific experts}
In MoE models, routing for English and for other languages differs most in the first and last layers, and the routing
of non-English text narrows more steeply with depth \citep{bandarkar2026multilingual}. Masking the experts that only one
language uses in Qwen3-30B-A3B damages mainly that language when done in early or late layers, and barely at all in
middle layers \citep{chen2026understanding}. Hebrew shows a ``deep-layer routing collapse'' in the same model
\citep{barjoseph2026mixture}, and language specialisation emerges in the final MoE layers
\citep{khandelwal2026leveraging}. In dense models, transfer neurons move representations into and out of a shared space
at the two ends of the network \citep{tezuka2025transfer}.

\paragraph{Lithuanian language technology}
\citet{nakvosas2025open} release Lithuanian Llama2 models and translated benchmarks, and TildeOpen
\citep{bergmanis2026tildeopen} is an open 30B model built to serve Baltic and other European languages fairly.
\citet{jakubauskaite2026evaluating} test LLMs on 305 minimal pairs of Lithuanian noun cases and find a monolingual
Lithuanian model ahead of multilingual ones. Closest to our question, \citet{kapociutedzikiene2025localizing} find that
4-bit quantisation costs Lithuanian more than English. We extend this evidence to MoE expert pruning, where the gap
turns out to be much larger.

\begin{table}[ht]
\caption{How this study differs from the closest prior work (``Calib.\ lang.'': calibration language varied;
``Gen.'': open generation evaluated).}
\label{tab:prior}
\footnotesize
\setlength{\tabcolsep}{3pt}
\begin{tabular}{@{}L{2.45cm}L{2.2cm}L{1.75cm}L{1.9cm}L{1.8cm}L{0.85cm}@{}}
\toprule
Work & Models & Compression & Calib.\ lang. & Languages & Gen. \\
\midrule
\citet{salim2026expert} & gpt-oss-20b (MoE) & expert pruning & EN vs ZH & EN, 7 in MMLU-ProX & no \\
\citet{salim2026towards} & gpt-oss-20b (MoE) & expert pruning & yes & 7 high-resource & partly \\
\citet{martin2026extracting} & gpt-oss-20b, Qwen3-30B (MoE) & expert pruning & translation data & 4 + 3 unseen & yes \\
\citet{chen2026understanding} & Qwen3-30B (MoE) & expert masking & -- & 10 & yes \\
\citet{choenni2025mwanda} & 6 dense & Wanda pruning & multilingual & 15 & no \\
\citet{chimoto2026calibrating} & 2 dense & quantisation & yes & 10 & no \\
\citet{kapociutedzikiene2025localizing} & dense & quantisation & -- & LT, LV, ET & yes \\
This work & 4 MoE & REAP pruning & EN/LT/MIX, 2 domains & EN, LT & yes \\
\bottomrule
\end{tabular}
\end{table}

\section{Experimental Setup}\label{sec:method}

\paragraph{REAP expert pruning}
REAP \citep{lasby2025reap} gives each routed expert $j$ of an MoE layer the saliency score
\begin{equation}\label{eq:reap}
S_j = \frac{1}{|\mathcal{X}_j|}\sum_{x\in\mathcal{X}_j} g_j(x)\,\lVert f_j(x)\rVert_2 ,
\end{equation}
where $\mathcal{X}_j$ is the set of calibration tokens routed to $j$, $g_j(x)$ is the router's gate weight and $f_j(x)$
the expert's output. In words, an expert matters if, on the tokens it receives, the router trusts it and its output is
large. Because the score is an average, an expert that few tokens reach can still score high, while an expert that no
calibration token reaches scores 0 and is removed first. In every MoE layer, REAP removes the lowest-scoring 25\% or
50\% of the routed experts (the pruning ratio); shared experts, dense layers and the number of active experts are
unchanged. We use the implementation in llm-compressor 0.15.0 \citep{redhat2026llm}, which prunes
one layer at a time; \appref{sec:impl} gives the details, including two patches needed for Gemma~4.

\begin{table}[ht]
\caption{The four models (a) and their unpruned scores (b). Tokens per word: mean tokens per word in each model's
FLORES translations into that language. Bits per byte: Wikipedia NLL per UTF-8 byte, in bits (lower is better).
$^{a}$Gemma~4 runs a dense MLP alongside its routed experts. $^{b}$GLM's first layer is dense. Parameter counts include
the vision towers of Gemma~4 and Qwen3.6 and exclude multi-token-prediction heads. Gemma~4's bits per byte are high
because it is an instruction-tuned model scored inside its chat template; compare it only with itself.}
\label{tab:models}
\footnotesize
\setlength{\tabcolsep}{3pt}
(a)\\[2pt]
\resizebox{\textwidth}{!}{\begin{tabular}{@{}llrrrrrrr@{}}
\toprule
 & & MoE & routed & & & params & \multicolumn{2}{c}{tokens/word} \\
\cmidrule(l){8-9}
Model & short name & layers & experts & top-$k$ & shared & (B) & EN & LT \\
\midrule
\texttt{Qwen3-30B-A3B-Instruct-2507} & Qwen3-30B & 48 & 128 & 8 & -- & 30.5 & 1.28 & 3.23 \\
\texttt{Qwen3.6-35B-A3B} & Qwen3.6 & 40 & 256 & 8 & 1 & 34.7 & 1.27 & 2.64 \\
\texttt{gemma-4-26B-A4B-it} & Gemma 4 & 30 & 128 & 8 & 1$^{a}$ & 25.8 & 1.25 & 2.56 \\
\texttt{GLM-4.7-Flash} & GLM-4.7-Flash & 46$^{b}$ & 64 & 4 & 1 & 29.9 & 1.25 & 3.14 \\
\bottomrule
\end{tabular}
}\\[6pt]
(b)\\[2pt]
\centering
\begin{tabular}{@{}lrrrrrr@{}}
\toprule
 & \multicolumn{2}{c}{Belebele acc.\ (\%)} & \multicolumn{2}{c}{FLORES chrF++} & \multicolumn{2}{c}{Wikipedia bits/byte} \\
\cmidrule(lr){2-3}\cmidrule(lr){4-5}\cmidrule(l){6-7}
Model & EN & LT & LT$\to$EN & EN$\to$LT & EN & LT \\
\midrule
Qwen3-30B & 94.3 & 89.9 & 57.8 & 47.5 & 0.71 & 1.05 \\
Qwen3.6 & 95.4 & 90.8 & 59.6 & 54.0 & 0.66 & 0.77 \\
Gemma 4 & 95.9 & 91.1 & 58.8 & 50.7 & 0.98 & 1.81 \\
GLM-4.7-Flash & 84.8 & 73.8 & 58.1 & 49.0 & 0.71 & 0.87 \\
\bottomrule
\end{tabular}

\end{table}

\paragraph{Models}
We prune four open MoE models (\tabref{tab:models}): Qwen3-30B-A3B-Instruct-2507
\citep{yang2025qwen3,qwen2025qwen3}, Qwen3.6-35B-A3B \citep{qwen2026qwen36,qwen2026qwen36blog}, gemma-4-26B-A4B-it
\citep{gemma2026gemma,google2026gemma} and GLM-4.7-Flash \citep{zai2026glm,glm2025glm45}, which we call Qwen3-30B,
Qwen3.6, Gemma~4 and GLM-4.7-Flash. They differ in the number of experts (64--256), active experts (4 or 8), shared
experts and router design. Before pruning, all four are weaker in Lithuanian than in English on all three benchmarks
(\tabref{tab:models}b). Pruning 25\% of the experts
removes \V{params-25-min}--\V{params-25-max}\% of the parameters, and pruning 50\% removes
\V{params-50-min}--\V{params-50-max}\% (\tabref{tab:compression}). Thinking mode is off throughout.

\paragraph{Calibration data}
Each calibration set has 1{,}024 samples of 2{,}048 tokens from one of two domains. \textbf{Wikipedia}
\citep{wikimedia2023wikipedia} provides running encyclopaedic text. \textbf{Europarl} \citep{koehn2005europarl} provides
parliamentary proceedings that exist in both languages, so that English and Lithuanian calibration see the same
content. Calibration is English only (\textbf{EN}), Lithuanian only (\textbf{LT}) or a 50/50 mix of the two
(\textbf{MIX}). The budget is fixed in tokens, and Lithuanian needs more tokens per word, so LT calibration covers about
half as much text as EN calibration. Each sample is wrapped as an assistant reply in the model's chat format, because
Gemma~4 works as a language model only in that format; saliency is computed over all tokens, including the short chat
prefix. Four models, two domains, three languages and two pruning ratios (25\% and 50\% of the routed experts in each
layer) give 48 pruned models. We name a configuration by its domain, language and ratio, e.g.\ \emph{Europarl-LT,
25\%}, and call a domain--language pair a \emph{calibration set}. \appref{sec:impl} gives the sampling details and the
exact prompts.

\paragraph{Evaluation}
We evaluate all 52 models (48 pruned and 4 unpruned) in both languages, always prompting in the language of the input.
\begin{itemize}
\item \textbf{Belebele} \citep{bandarkar2024belebele}: the same 900 reading-comprehension questions in each language.
The answer is whichever of the letters A--D the model finds most probable; chance is 25\%. Besides this ordinary
accuracy, we report a \emph{strict} accuracy that counts an answer only if its letter has a probability above 0.5, and
the distribution of predicted letters.
\item \textbf{FLORES-200} \citep{nllb2022language,nllb2024scaling,goyal2022flores}: the 1{,}012 devtest sentences
translated from English into Lithuanian (EN$\to$LT) and back (LT$\to$EN), with greedy decoding and at most 256 new
tokens. We score the raw outputs with chrF++ \citep{popovic2017chrf,post2018call} and BLEU \citep{papineni2002bleu}.
\item \textbf{Wikipedia}: the token negative log-likelihood (NLL) of 5{,}000 held-out articles per language, wrapped in
the same chat prompt as the calibration data. The same runs give the routing statistics of Section~\ref{sec:rq5}.
\end{itemize}

\paragraph{How we report results}\label{sec:conventions}
For a language $\ell$, the \emph{degradation} $D_\ell$ and the LT--EN \emph{gap} $G$ are
\begin{equation}\label{eq:gap}
D_\ell = s_\ell^{\mathrm{unpruned}} - s_\ell^{\mathrm{pruned}}, \qquad G = D_{\mathrm{LT}} - D_{\mathrm{EN}},
\end{equation}
where $s$ is Belebele accuracy (\%) or chrF++; for Wikipedia, $D_\ell$ is the relative NLL increase in \%,
${100\,(\mathrm{NLL}_\ell^{\mathrm{pruned}}/\mathrm{NLL}_\ell^{\mathrm{unpruned}} - 1)}$. A positive gap means that
Lithuanian lost more. For translation, $D_{\mathrm{LT}}$ uses EN$\to$LT (Lithuanian output) and $D_{\mathrm{EN}}$ uses
LT$\to$EN. Gaps are in percentage points (pp), except for FLORES (chrF++ points). We never compare perplexities across
languages, as Lithuanian is split into more tokens. When we compare two calibration choices, a positive
\emph{effect} means that the first is better (for Wikipedia, lower NLL). A run has \emph{collapsed} when its Belebele
accuracy is at most 30\%.

\paragraph{Statistics}
Each calibration language has 16 configurations (4 models $\times$ 2 domains $\times$ 2 ratios), 8 at each ratio, and
each ratio has 24 (4 models $\times$ 6 calibration sets). We give 95\% confidence intervals from paired bootstrap
resampling of the test items \citep{efron1993introduction,koehn2004statistical}; for Belebele and FLORES the parallel
items of the two languages are resampled together. A difference is \emph{significant} when its interval excludes 0,
and ``14/16'' means that 14 of 16 configurations show a significant effect. These per-configuration tests are not
independent replications, because the configurations share four base models and the same test items, so we also
report means per model. Our main endpoints are the Belebele and FLORES gaps and the effects of LT versus EN
calibration; the other analyses are descriptive. Unless stated otherwise, means are over the four models and the two
domains.

\section{Results}\label{sec:results}

\subsection{RQ1: Does REAP degrade Lithuanian more than English?}\label{sec:rq1}

\begin{table}[t]
\caption{Mean LT--EN gap over the four models and two domains (positive = Lithuanian loses more), and the number of the
8 configurations whose gap is significantly positive/negative.}
\label{tab:gaps}
\small
\setlength{\tabcolsep}{4pt}
\centering
\begin{tabular}{@{}llrcrcrc@{}}
\toprule
 & & \multicolumn{2}{c}{Belebele (pp)} & \multicolumn{2}{c}{FLORES (chrF++)} & \multicolumn{2}{c}{Wikipedia (pp)} \\
\cmidrule(lr){3-4}\cmidrule(lr){5-6}\cmidrule(lr){7-8}
Calib. & Ratio & gap & sig.\,$+$/$-$ & gap & sig.\,$+$/$-$ & gap & sig.\,$+$/$-$ \\
\midrule
EN & 25\% & +23.5 & 8/0 & +12.8 & 7/0 & +50.0 & 8/0 \\
 & 50\% & +30.6 & 6/0 & +8.5 & 5/2 & +116.7 & 8/0 \\
\addlinespace[2pt]
LT & 25\% & +3.0 & 5/0 & $-$0.3 & 0/3 & $-$2.4 & 0/8 \\
 & 50\% & +5.1 & 6/1 & +1.2 & 2/3 & $-$8.1 & 0/8 \\
\addlinespace[2pt]
MIX & 25\% & +5.6 & 5/0 & +0.1 & 1/1 & $-$1.2 & 0/7 \\
 & 50\% & +6.2 & 5/0 & $-$0.1 & 2/2 & $-$2.6 & 3/4 \\
\bottomrule
\end{tabular}

\end{table}

\begin{figure}[t]
\includegraphics[width=\textwidth]{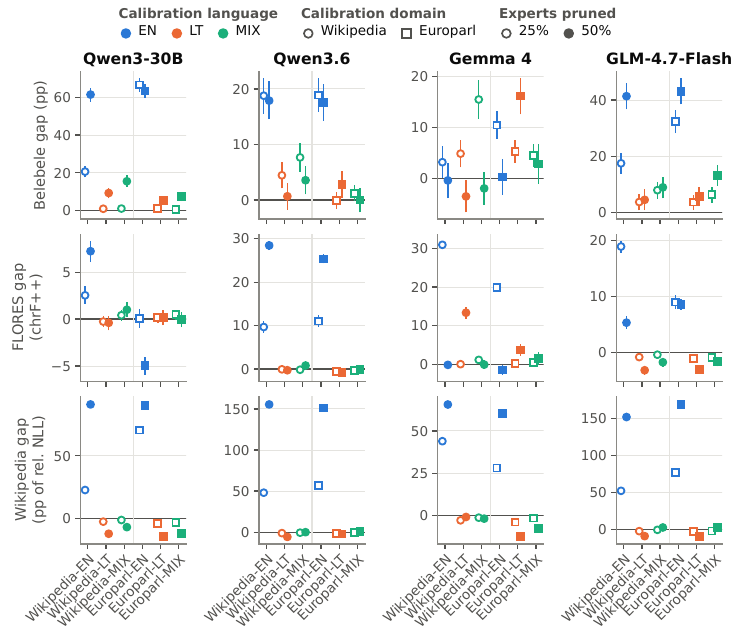}
\caption{\textbf{LT--EN gap with 95\% intervals} for each model (columns) and calibration set (x axis). Rows: Belebele
(pp), FLORES (chrF++) and Wikipedia (pp of relative NLL increase), with different y scales. Positive means that
Lithuanian loses more; markers as in \figref{fig:headline}. English-calibrated gaps are large except for Gemma~4 and
runs where both languages collapsed; LT and MIX gaps are small except for Gemma~4.}
\label{fig:ltengap}
\end{figure}

\textbf{Yes, and by a wide margin under English calibration} (\tabref{tab:gaps}, \figref{fig:ltengap}). At 25\%
pruning, English-calibrated models lose a median of \V{encal25-med-en} Belebele points in English but
\V{encal25-med-lt} in Lithuanian. Their translation into Lithuanian loses \V{gap-flo-en-25-u} chrF++ more than their
translation into English, and their Lithuanian language modelling degrades far more than their English. The gap is
positive for every model on every benchmark and, over both ratios, significant in at least \V{sigpos-en-min} of the 16
English-calibrated configurations on each benchmark. This supports H1 for English calibration.

One example shows the pattern. Qwen3-30B pruned by 25\% with \emph{Europarl-EN} keeps \V{q3euen25-en}\% accuracy on
English Belebele, only \V{q3euen25-en-loss} point below the unpruned model, while its Lithuanian accuracy falls from
\V{q3euen25-lt-base}\% to \V{q3euen25-lt}\%, no better than guessing. English stays nearly intact in three of the four
models; the exception is Gemma~4, which with \emph{Wikipedia-EN} also loses \V{gwien25-en-loss} points in English.
Strict scoring widens the contrast: the English-calibrated models at 25\% keep \V{strict-en-25-en}\% strict accuracy in
English, almost all of their ordinary accuracy, but only \V{strict-en-25-lt}\% in Lithuanian, where their ordinary
accuracy is \V{acc-en-25-lt}\%.

\textbf{With Lithuanian or mixed calibration, a small gap remains.} On Belebele it averages
\V{gap-bel-lt-25-u}--\V{gap-bel-lt-50-u}\pp{} under LT and \V{gap-bel-mix-25-u}--\V{gap-bel-mix-50-u}\pp{} under MIX
calibration. It is not explained by Lithuanian starting from a lower score: on questions that both languages answered
correctly before pruning, LT-calibrated models at 25\% lose more Lithuanian than English answers in all 8
configurations. The remaining gap is smallest for Qwen3.6, whose English and Lithuanian saliency rankings agree most.
On FLORES the gap is close to zero, and on Wikipedia Lithuanian even degrades less than English.

\textbf{A gap near zero can also mean that both languages collapsed.} Gemma~4 at 50\% answers Belebele at chance in
Lithuanian with every calibration set, and in English with \V{g50-en-chance-n} of the 6. Collapsed runs explain the
only significantly negative Belebele gap (Gemma~4, \emph{Wikipedia-LT, 50\%}) and both significantly negative FLORES
gaps under English calibration (\figref{fig:gapheatmap}). \appref{sec:full} gives further statistics.

\subsection{RQ2: Does the calibration language change the gap?}\label{sec:rq2}

\begin{table}[t]
\caption{Effect of switching the calibration language: mean difference over 16 configurations (positive = the first is
better; Wikipedia: NLL reduction in nats per token). B/W: how many are significantly better/worse.}
\label{tab:langeffect}
\small
\setlength{\tabcolsep}{3.5pt}
\centering
\begin{tabular}{@{}lrcrcrc@{}}
\toprule
 & \multicolumn{2}{c}{LT $-$ EN} & \multicolumn{2}{c}{MIX $-$ EN} & \multicolumn{2}{c}{MIX $-$ LT} \\
\cmidrule(lr){2-3}\cmidrule(lr){4-5}\cmidrule(lr){6-7}
Evaluation & mean & B/W & mean & B/W & mean & B/W \\
\midrule
Belebele EN (pp) & +2.5 & 6/0 & +1.4 & 7/1 & $-$1.1 & 4/4 \\
Belebele LT (pp) & +25.5 & 15/0 & +22.5 & 14/1 & $-$3.0 & 0/7 \\
FLORES LT$\to$EN (chrF++) & +18.0 & 16/0 & +16.5 & 16/0 & $-$1.5 & 3/2 \\
FLORES EN$\to$LT (chrF++) & +28.2 & 16/0 & +27.2 & 16/0 & $-$1.0 & 1/8 \\
Wikipedia EN (nats) & $-$0.06 & 5/11 & $-$0.03 & 5/10 & +0.03 & 12/3 \\
Wikipedia LT (nats) & +1.57 & 16/0 & +1.54 & 16/0 & $-$0.03 & 2/14 \\
\bottomrule
\end{tabular}

\end{table}

\begin{figure}[t]
\includegraphics[width=\textwidth]{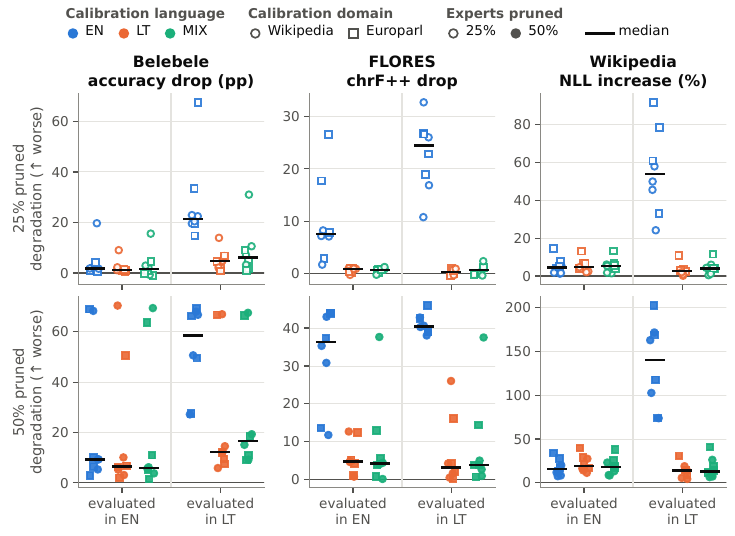}
\caption{\textbf{Transfer between calibration languages is asymmetric.} Degradation (positive = worse) of English
(left) and Lithuanian (right) for each calibration language, at 25\% (top) and 50\% (bottom); one point per model and
domain; black bars are medians. For FLORES, English is LT$\to$EN. Every calibration language preserves English; only
LT and MIX preserve Lithuanian.}
\label{fig:transfer}
\end{figure}

\textbf{The calibration language is the factor that matters most for the gap} (\tabref{tab:langeffect},
\figref{fig:transfer}). On its own (one-way $\eta^2$), it explains \V{eta-belgap-lang}\% of the variation of the
Belebele gap across the 48 configurations, \V{eta-flogap-lang}\% for FLORES and \V{eta-wikigap-lang}\% for Wikipedia,
far more than the model, the pruning ratio or the calibration domain (\tabref{tab:eta}). For Lithuanian accuracy
itself, the model and the ratio matter about as much as the language.

Switching from English to Lithuanian calibration improves Lithuanian Belebele accuracy by \V{eff-lten-bel-lt-u} points
on average and translation into Lithuanian by \V{eff-lten-enlt-u} chrF++ (\figref{fig:forestlten}). The gain appears in
every model and almost every configuration; the one exception, Gemma~4 at 50\% with Wikipedia calibration, is at chance
either way. This supports H2. Two checks rule out simple explanations. First, the gain is larger on Europarl, where
both languages see the same content (\V{ltben-europarl-u} points, against \V{ltben-wikipedia-u} with Wikipedia), so it
does not come from differences between the English and Lithuanian Wikipedias. Second, it is not limited to Lithuanian
output: translation into English, where Lithuanian is only read, improves by \V{eff-lten-lten-u} chrF++.

\textbf{English barely changes, so the transfer is asymmetric} (\figref{fig:transfer}). Under LT calibration, English
Belebele is never significantly worse than under EN calibration and is on average \V{eff-lten-bel-en-u} points better,
so H3 holds for comprehension; under MIX calibration it is significantly worse in \V{eff-mixen-bel-en-worse} of the 16
configurations. Language modelling shows the asymmetry most clearly. At 25\%, Lithuanian NLL rises by
\V{relnll-lt-25-lt}\% under LT calibration but by \V{relnll-en-25-lt}\% under EN calibration, whereas English NLL rises
by \V{relnll-en-25-en}\% under EN and \V{relnll-lt-25-en}\% under LT calibration; at 50\%, the English cost of LT
calibration is larger but still modest (\appref{sec:effects}). With LT calibration at 25\%, translation stays within
\V{ltcal25-flo-maxdrop} chrF++ of the unpruned model in every model and direction. We did not evaluate
open-ended English generation. Section~\ref{sec:rq5} looks at why the transfer is one-sided.

\textbf{A mix works almost as well as Lithuanian alone} (\tabref{tab:langeffect}). MIX
calibration is slightly worse than LT for Lithuanian (\V{eff-mixlt-bel-lt-u} points on Belebele and
\V{eff-mixlt-enlt-u} chrF++ into Lithuanian) and slightly better for English language modelling. It does not balance
the two languages better, however: compared with LT calibration, it lowers the larger of a configuration's English and
Lithuanian Belebele losses in only \V{h5-worst-mix-better} of the 16 configurations, so H4 is not supported.

\subsection{RQ3: How does the gap depend on the pruning ratio?}\label{sec:rq3}

\begin{figure}[t]
\includegraphics[width=\textwidth]{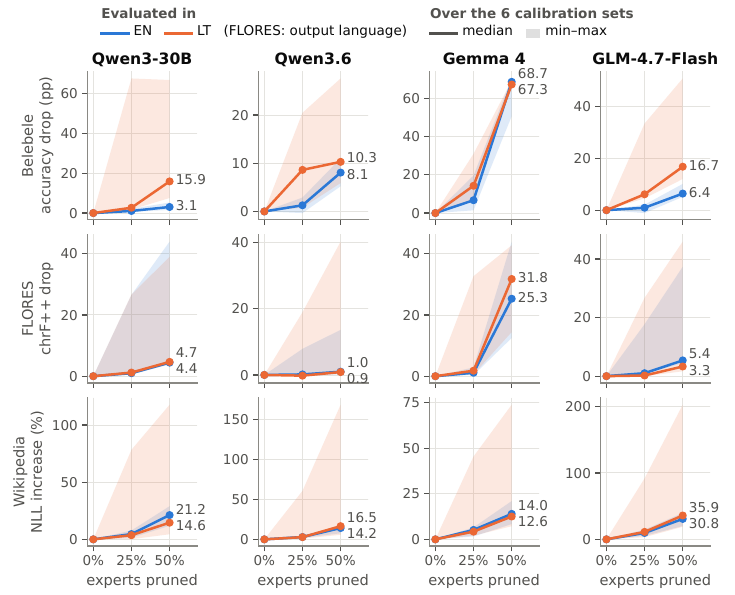}
\caption{\textbf{Degradation at 0, 25 and 50\% of experts pruned} (positive = worse). Lines are medians over the six
calibration sets and bands span their minimum to maximum; colour is the evaluated language (FLORES: output language).
Losses grow moderately up to 25\% and steeply to 50\%; the worst Lithuanian cases are usually English-calibrated.
Gemma~4 collapses at 50\%.}
\label{fig:ratiotrend}
\end{figure}

\textbf{Pruning half of the experts does far more damage than pruning a quarter} (\figref{fig:ratiotrend}). Going from 25\% to 50\% makes at least \V{ratio-worse-min} of the 24 configurations
significantly worse on every benchmark, and the average losses become \V{ratio-loss-x-min}--\V{ratio-loss-x-max} times
larger. Even with Lithuanian calibration, the models other than Gemma~4, which collapses, lose
\V{ltcal50-nogemma-lt-loss-min}--\V{ltcal50-nogemma-lt-loss-max} points of Lithuanian Belebele accuracy at 50\%.

\textbf{The gap grows consistently only for language modelling.} Under English calibration, the Wikipedia gap grows from
25\% to 50\% in all \V{h3-wikipedia-en-up} configurations, whereas the Belebele and FLORES gaps grow in some
configurations and shrink in others (\tabref{tab:h3}), although the mean Belebele gap rises (\tabref{tab:gaps}). These
two gaps are capped: Lithuanian Belebele cannot fall below chance, which Qwen3-30B with \emph{Europarl-EN} already
reaches at 25\%, and at 50\% English translation breaks as well. H5 is therefore supported only for Wikipedia under
English calibration.

\subsection{RQ4: Does the calibration domain matter?}\label{sec:rq4}

\textbf{The calibration domain matters much less than the language} (\appref{sec:effects}). Its main effects are
small: Wikipedia calibration lowers NLL on the in-domain Wikipedia test articles in most configurations, and
Europarl calibration helps English Belebele by \V{ds-bel-en-mean-u} points on average. On its own, the domain explains
almost none of the variation of the gap (\tabref{tab:eta}). The exception is Lithuanian under English calibration, where
the domain can swing accuracy widely: Qwen3-30B at 25\% scores \V{q3euen25-lt}\% on Lithuanian Belebele with
\emph{Europarl-EN} but \V{q3wien25-lt}\% with \emph{Wikipedia-EN}. A possible reason, which we did not test, is that
English Wikipedia contains names and other non-English material that keep a few Lithuanian-relevant experts in use.

\subsection{RQ5: Are the experts that matter language-specific?}\label{sec:rq5}

\textbf{Partly. The two languages share most experts, and the ones they do not share matter for Lithuanian.} English
and Lithuanian calibration rank experts similarly but not identically: the Spearman correlation of their saliency
scores is \V{sal-enlt-min}--\V{sal-enlt-max} on average over layers, depending on the model, with a few layers near
zero (\figref{fig:sallayers}). With the same domain, the two languages therefore keep
\V{shared-enlt-25-min}--\V{shared-enlt-25-max}\% of the same experts at 25\% pruning, and Lithuanian text is routed to
a narrower set of experts than English text in every model (\appref{sec:extra}). In three of the four models, a change
of calibration domain changes the kept set at least as much as a change of language (\figref{fig:overlappairs}),
although the language has by far the larger effect on quality: which experts are removed matters more than how many
differ.

\begin{figure}[t]
\includegraphics[width=\textwidth]{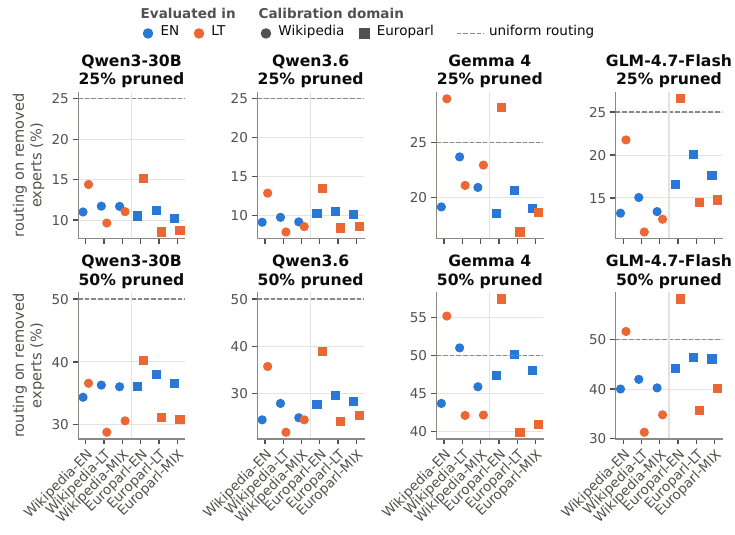}
\caption{\textbf{Removed routing share}: the share of each language's routing that went to the experts pruning removed,
measured on the unpruned model for English (blue) and Lithuanian (orange) Wikipedia text, at 25\% (top) and 50\%
(bottom); circles and squares are Wikipedia and Europarl calibration; the dashed line is uniform routing. Under EN
calibration Lithuanian loses more routing than English in every model; under LT calibration the order reverses.}
\label{fig:removedmass}
\end{figure}

\textbf{The experts removed under English calibration had carried more Lithuanian routing} (\figref{fig:removedmass}).
To see this, we take the unpruned model's expert selections on the Wikipedia test text and compute, for each language,
the share that went to the experts that pruning later removes (the \emph{removed routing share}). Under English
calibration at 25\%, the removed experts had received \V{remmass-en-25-lt}\% of the Lithuanian selections but only
\V{remmass-en-25-en}\% of the English ones. Lithuanian calibration reverses the order (\V{remmass-lt-25-lt}\% against
\V{remmass-lt-25-en}\%), in every model and at both ratios. Across all 24 configurations at 25\%, the larger the
Lithuanian share removed, the more Lithuanian accuracy is lost (Spearman $\rho=\V{r12-all_25-lt-ltdeg}$).

\textbf{Lithuanian also suffers more from each unit of routing lost} (\figref{fig:coverage}). Leaving out Gemma~4, each
percentage point of Lithuanian routing removed goes with about \V{cov-slope-lt-u} points of lost Lithuanian Belebele
accuracy, whereas removing English routing goes with no measurable loss in English. The asymmetric transfer of
Section~\ref{sec:rq2} therefore has two sources: which experts the calibration protects, and how much each language
depends on the experts it loses.

\begin{figure}[t]
\includegraphics[width=\textwidth]{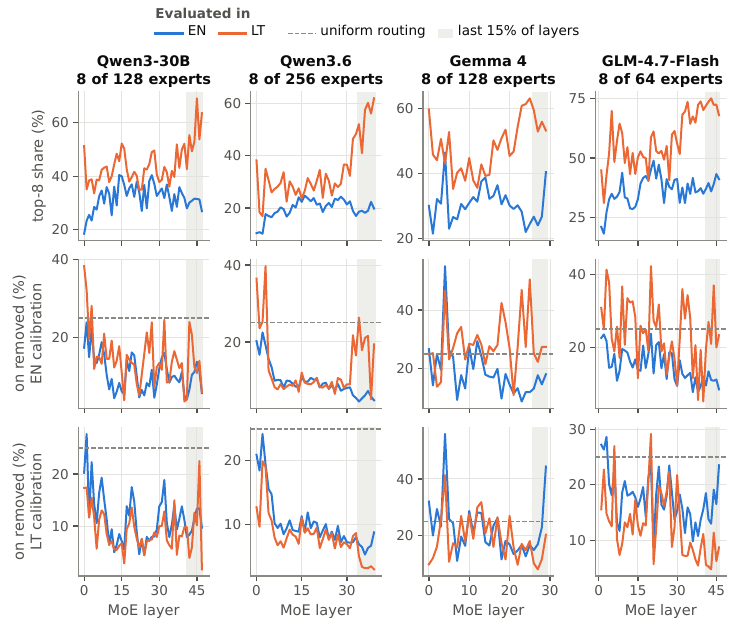}
\caption{\textbf{Lithuanian relies on few experts at the end of the network, and EN calibration removes some of them.}
Top: share of routing handled by the 8 most-used experts in each MoE layer (unpruned; colour = evaluated language).
Middle and bottom: removed routing share of each language at 25\% under EN and LT calibration (mean of the two
domains); dashed = uniform; the last 15\% of layers are shaded. How much more Lithuanian routing EN calibration removes
there depends on the model.}
\label{fig:routingdepth}
\end{figure}

\textbf{The languages differ most at the two ends of the network} (\figref{fig:routingdepth}, \tabref{tab:depth}). In
the middle layers English and Lithuanian use the experts in a similar way; in the first and especially the last layers
their routing diverges, as \citet{bandarkar2026multilingual} found for other languages. At the output end, Lithuanian
leans on a few experts: in the last layers the 8 most used handle \V{depth-last-top-lt-min}--\V{depth-last-top-lt-max}\%
of its expert selections, against \V{depth-last-top-en-min}--\V{depth-last-top-en-max}\% for English, and English
calibration removes on average \V{top8-last-encal-lt-min}--\V{top8-last-encal-lt-max} of these 8 experts per layer,
depending on the model (Lithuanian calibration at most \V{top8-last-ltcal-lt-max}). At the input end, the Lithuanian
routing removed in the first layers tracks the loss of accuracy most closely ($\rho=\V{r12-all_25-lt_first-ltdeg}$), so
the input side matters at least as much as the output side. Where in depth the extra Lithuanian routing is removed
varies between models (\appref{sec:extra}).

\textbf{Removed routing shows which calibration is worse, not how bad the damage will be.} Across calibration
languages, the Lithuanian share removed tracks the loss, but within English calibration the extra Lithuanian routing
removed (the Lithuanian minus the English share) does not predict the size of the gap: Gemma~4 with \emph{Europarl-EN,
25\%} loses the largest share of Lithuanian routing but less accuracy than Qwen3-30B. Neither the overlap of the kept
experts nor the saliency correlation predicts the gaps (\appref{sec:extra}).

\subsection{RQ6: How do pruned models fail?}\label{sec:rq6}

\textbf{English-calibrated models often do not stop when writing Lithuanian.} GLM-4.7-Flash, for example, runs into the
256-token limit in \V{hits-glm-en25-min}--\V{hits-glm-en25-max}\% of its EN$\to$LT translations at 25\%
(\appref{sec:failures}). At 50\% the models also switch language: \V{off-en-50-enlt}\% of the EN$\to$LT outputs of
English-calibrated models contain more English than Lithuanian words, against \V{off-lt-50-enlt}\% with Lithuanian
calibration and none for the unpruned models (\appref{sec:failures}).

\begin{figure}[t]
\includegraphics[width=\textwidth]{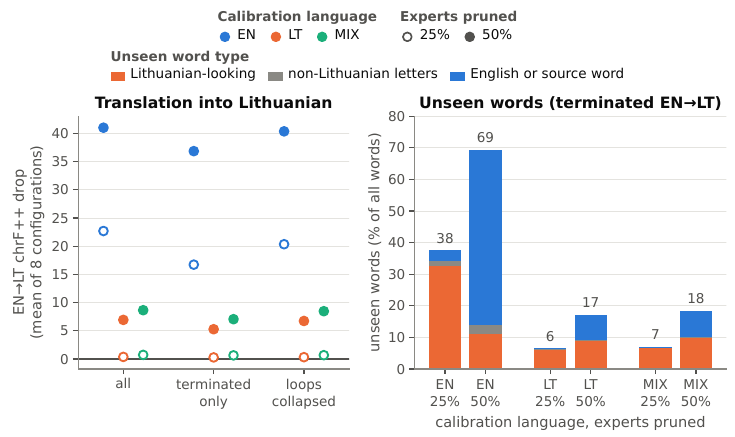}
\caption{\textbf{Lithuanian translation breaks beyond runaway generation.} Left: mean EN$\to$LT chrF++ drop (positive =
worse) on all sentences, on the sentences the model finished (compared with the unpruned model on the same sentences)
and with repetition loops collapsed; markers as in \figref{fig:headline}. Right: words of finished EN$\to$LT
translations that occur in no reference or unpruned output (\% of all words), split into Lithuanian-looking forms
(orange), words with non-Lithuanian letters (grey) and English or source words (blue). At 25\% EN calibration mostly
produces unattested Lithuanian-looking forms, at 50\% mostly English.}
\label{fig:floresquality}
\end{figure}

\textbf{The content breaks as well} (\figref{fig:floresquality}). Runaway outputs explain only about a quarter of the
translation loss: under English calibration at 25\%, EN$\to$LT loses \V{a4-en-25-enlt-all} chrF++ overall and still
\V{a4-en-25-enlt-kept} on the sentences the model finished. In these finished translations, about a third of the words
(\V{split-en-25-ltlike}\%) look Lithuanian but occur in no reference or unpruned output, against \V{split-lt-25-ltlike}\%
under Lithuanian calibration; this measure cannot tell a rare correct form from an invented one. At 50\%, English or
source words take over (\V{split-en-50-eng}\% of all words).

\textbf{Typical Lithuanian errors.} The pruned models produce (i)~non-words and wrong stems where the unpruned model was
correct, such as ``ai\v{s}\v{c}iai'' for \emph{ai\v{s}kiai} (`clearly') and ``pristat\.{e}jo'' for \emph{pristat\.{e}}
(`presented') from Qwen3-30B with \emph{Europarl-EN, 25\%}; (ii)~broken diacritics, where the model emits only
part of the UTF-8 bytes of a Lithuanian letter, which then decode to an invalid character (\V{fffd-total} outputs,
\V{fffd-glm-pct}\% of them from GLM-4.7-Flash and none from unpruned models); (iii)~switches into English; and (iv)~repetition loops.

\textbf{Collapsed models answer ``A''.} When Belebele accuracy collapses, a model puts most of its answers on one
letter, usually ``A'' (\figref{fig:collapse}); Gemma~4 at 50\%, for example, answers ``A''
to \V{g50-A-min}--\V{g50-A-max}\% of the questions. Pruned Gemma~4 also writes reasoning text although thinking is
switched off; this text is scored as part of its translations, but removing it changes the results only modestly.
\appref{sec:failures} gives the details and two checks of the metrics.

\section{Discussion}\label{sec:discussion}

\tabref{tab:hyp} summarises the five hypotheses and their verdicts.

\begin{table}[t]
\caption{Hypotheses, decision rules and verdicts. Counts are configurations with a significant effect.}
\label{tab:hyp}
\footnotesize
\begin{tabular}{@{}L{2.5cm}L{3.7cm}L{5.0cm}@{}}
\toprule
Hypothesis & Decision rule & Verdict \\
\midrule
H1: pruning degrades LT more than EN & gap significantly positive in most of the 16 configurations of a calibration
language & \textbf{Supported under EN calibration} (\V{sigpos-bel-en}/16 Belebele, \V{sigpos-flo-en}/16 FLORES,
\V{sigpos-wiki-en}/16 Wikipedia); under LT/MIX only a small Belebele gap (\V{sigpos-bel-lt}/16, \V{sigpos-bel-mix}/16) \\
H2: LT calibration reduces LT degradation & LT better on a Lithuanian evaluation in most of the 16 configurations and
on average in all 4 models & \textbf{Supported}: \V{eff-lten-enlt-better}/16 EN$\to$LT, \V{eff-lten-wiki-lt-better}/16
Wikipedia, \V{eff-lten-bel-lt-better}/16 Belebele; all four models \\
H3: LT/MIX preserve LT without hurting EN & English Belebele never significantly worse than under EN &
\textbf{Supported under LT} for comprehension; \V{eff-mixen-bel-en-worse}/16 worse under MIX; English NLL rises
slightly \\
H4: MIX more balanced than LT & MIX reduces the worse language's Belebele loss in most of the 16 configurations &
\textbf{Not supported} (\V{h5-worst-mix-better}/16) \\
H5: the gap grows from 25\% to 50\% & change in gap significantly positive in most of 8 configurations &
\textbf{Only for Wikipedia under EN calibration} (\V{h3-wikipedia-en-up}/8); Belebele (\V{h3-belebele-en-up}/8) and
FLORES (\V{h3-flores-en-up}/8) capped by chance and English collapse \\
\bottomrule
\end{tabular}
\end{table}

\paragraph{Why English calibration hurts Lithuanian}
Our results suggest a simple account, although the evidence for it is correlational (Section~\ref{sec:rq5}). English
and Lithuanian use largely the same experts in the middle of the network, as expected from the shared middle layers of
multilingual models \citep{tezuka2025transfer,bandarkar2026multilingual}, but at the two ends Lithuanian takes its own
routes, and in the last layers it relies on a few experts, as \citet{barjoseph2026mixture} report for Hebrew. REAP
judges an expert only by the calibration tokens routed to it (Eq.~\ref{eq:reap}). Under English calibration, an expert
that Lithuanian relies on sees only the English tokens that happen to reach it, where it is usually a minor choice, or
no tokens at all; it therefore looks unimportant and is removed. Generation into Lithuanian suffers most, in line with
expert-masking results for other languages \citep{chen2026understanding}. Lithuanian calibration may transfer to
English because English routing is spread over many experts that Lithuanian text also uses in the shared middle layers.
The account does not fit every model equally well: Qwen3-30B collapses although the extra Lithuanian routing it loses
is small and spread evenly over depth. And the experts that look ``Lithuanian'' may encode sub-word or general
non-English features rather than Lithuanian as such.

\paragraph{Differences between models}
All four models show the main effects, but they differ in robustness. Qwen3.6, which has the most experts (256), a
shared expert and the closest agreement between English and Lithuanian saliency, is the most robust at 50\%: averaged
over all calibration sets, it loses \V{a6-all50-qsix} points of Belebele accuracy, against
\V{a6-all50-qthree}--\V{a6-all50-gemma} for the other models. Qwen3-30B and GLM-4.7-Flash hold up well under Lithuanian
calibration but are the most fragile under English calibration. Gemma~4 degrades in both languages already at 25\% and
collapses at 50\%. The number of tokens per word does not explain these differences: Qwen3.6 and Gemma~4 need the
fewest tokens per Lithuanian word, yet they are the most and the least robust models.

\paragraph{Recommendations for Lithuanian deployments}
Our results suggest three practices.
\begin{enumerate}
\item \textbf{Calibrate in Lithuanian or on a mix.} In our grid this also preserved English comprehension; a mix is the
safer choice when English language modelling matters most.
\item \textbf{Prune conservatively.} Losses grow steeply between 25\% and 50\%, even with Lithuanian calibration
(Section~\ref{sec:rq3}).
\item \textbf{Test in Lithuanian before deployment}, including generation, checks for degenerate output (runaway
length, wrong output language, letter collapse, reasoning text) and, where possible, a morphological test such as case
minimal pairs \citep{jakubauskaite2026evaluating}.
\end{enumerate}
The best compromise in our grid was Qwen3.6 pruned by 25\% with \emph{Europarl-LT} or \emph{Europarl-MIX} calibration:
it stayed within one point of the unpruned model on Belebele in both languages and within 0.2 chrF++ in both
translation directions.

\paragraph{Implications for evaluation}
An English-only evaluation would have missed the damage: at 25\%, \V{a5-encal25-en-within2} of the 8 English-calibrated
configurations stay within 2 points of their unpruned English accuracy while losing
\V{encal25-within2-lt-min}--\V{encal25-within2-lt-max} points in Lithuanian. It would not have misled the choice of
calibration set, because LT and MIX sets were also the best for English in \V{a5-ltmix-chosen} of the \V{a5-n}
model--ratio combinations (\tabref{tab:selection}), but it would have misreported the quality of the compressed
models. Multiple-choice scores also understate the damage to generation \citep{wen2026benchmark}: ordinary accuracy,
which takes the most probable letter, hides the loss of confidence that strict scoring reveals, and Belebele scores
near chance should be read together with the distribution of answer letters. Compression results should therefore be
reported per language and per task type, together with the calibration language.

\section{Limitations}\label{sec:limitations}

\begin{itemize}
\item \textbf{Design.} Each configuration was run once with greedy decoding, so our intervals reflect the sampling of
test items, not variation between calibration samples. Two ratios and one calibration budget cannot show where each
model starts to break.
\item \textbf{Data.} The fixed token budget gives Lithuanian about half as much calibration text as English, and the
256-token generation limit allows fewer than half as many Lithuanian words. The Wikipedia test articles are not
parallel (median scored length \V{wiki-median-bytes-lt} bytes in Lithuanian and \V{wiki-median-bytes-en} in English), which is why Europarl
serves as our content-controlled comparison. FLORES and Belebele may be in the models' training data.
\item \textbf{Scope.} We study one lower-resource language, and others may behave differently. We use llm-compressor's
REAP implementation, whose scores include the chat-prefix tokens, so an expert used mainly by the prefix can look
important: up to \V{prefix-sized-pct}\% of the experts on which EN and LT calibration disagree at 25\% have selection
counts small enough to come entirely from the prefix.
\item \textbf{Evaluation.} Belebele uses letter scoring (we add a strict variant, but no generative comprehension test).
Wikipedia NLL is computed inside the chat template and is not comparable with raw-text perplexity. We did not evaluate
open-ended English generation, instruction following or chat quality, and FLORES outputs were not post-processed.
\item \textbf{Analysis.} The link between removed routing and damage is correlational, and the overlap analysis has only
16 points from four models.
\end{itemize}
Natural next steps are to restore or ablate the experts that Lithuanian loses, to try non-uniform layer budgets,
calibration-free criteria \citep{liu2026aimer} and merging \citep{jha2026ream}, and to cover more ratios and languages,
with error annotation by native speakers and a morphological benchmark of the pruned models.

\section{Conclusion}\label{sec:conclusion}
REAP expert pruning is not language-neutral. Calibrated on English, it removes experts that Lithuanian relies on more
than English does, and Lithuanian comprehension, translation and language modelling degrade sharply at a pruning ratio
where English comprehension is nearly unchanged. Calibrating on Lithuanian, or on a 50/50 mix, removes most of this
damage at little cost to English. Although the two languages keep most of the same experts, the loss of Lithuanian
accuracy tracks how much Lithuanian routing the removed experts had carried. For Lithuanian, and plausibly for other
lower-resource languages, the calibration language should be chosen, and the compressed model evaluated, language by
language.

\FloatBarrier

\section*{Data and code availability}
The per-item outputs of all 52 models, the analysis tables and the scripts that regenerate every number, table and
figure of this paper are available from the corresponding author on reasonable request. The pruned models derive from
openly released checkpoints and are subject to their licences; the calibration and evaluation data are Wikipedia
(CC BY-SA), Belebele and FLORES-200 (both CC BY-SA 4.0), and Europarl.

\section*{Competing interests}
The authors are employees of Hostinger, which uses language models in its products. The authors declare no other
competing interests.

\section*{Use of generative AI}
A generative AI assistant (Claude, Anthropic) was used to help write and edit the text of this manuscript, improving
its language, clarity and structure. The authors designed and ran the experiments, reviewed and edited all generated
text, and take full responsibility for the content of the paper.

\bibliographystyle{plainnat}
\bibliography{references}

\begin{appendix}
\sloppy
\section*{Appendices}

The appendix uses the conventions of Section~\ref{sec:conventions}. In the effect forest (Figure~\ref{fig:forestlten}) a filled point means that
the 95\% interval excludes 0; elsewhere hollow = 25\% and filled = 50\% pruning. In tables, bold marks an
interval that excludes 0.
\FloatBarrier
\section{Implementation and data details}\label{sec:impl}

\textbf{Data.} Wikipedia: the \texttt{wikimedia/\allowbreak wikipedia} dumps \texttt{20231101.en} and \texttt{20231101.lt}; articles
shorter than 200 characters are dropped, the rest shuffled (seed 42); the first 5{,}000 articles form the evaluation
set and the next 20{,}000 the calibration pool, so the two never overlap. Europarl: \texttt{Helsinki-NLP/\allowbreak europarl},
pair \texttt{en-lt}; empty pairs are dropped, the rest shuffled (seed 42) and the first 150{,}000 pairs kept as the
calibration pool. EN and LT calibration take the English and the Lithuanian side of this pool in order, so the LT set
is the Lithuanian side of a subset of the pairs used for EN; the two halves of Europarl-MIX come from disjoint halves
of the pool. Wikipedia-MIX draws half of its samples from each language's pool. Belebele: \texttt{facebook/\allowbreak belebele} (\texttt{eng\_Latn},
\texttt{lit\_Latn}); FLORES-200: \texttt{facebook/\allowbreak flores} devtest. Calibration samples are not shuffled by the pruning
tool, and the 25\% and 50\% runs of a configuration use identical samples.

\textbf{Prompts.} Calibration samples and Wikipedia evaluation documents are the assistant turn of a chat whose user
turn is ``Write a passage of text.'' (EN) or ``Para\v{s}ykite teksto i\v{s}trauk\k{a}.'' (LT); the prefix before the
text is 14/20 tokens (EN/LT) for Qwen3-30B, 18/23 for Qwen3.6, 19/24 for Gemma~4 and 11/17 for GLM-4.7-Flash, identical
in calibration and evaluation. Belebele (user turn, in the question's language): ``Read the passage and answer the
question. Answer with only the letter of the correct option (A, B, C or D).'' followed by the passage, question and
options; Lithuanian: ``Perskaitykite tekst\k{a} ir atsakykite \k{i} klausim\k{a}. Atsakykite tik teisingo atsakymo raide
(A, B, C arba D).'' Each letter is scored as the summed probability of its single-token forms with and without a
leading space. FLORES (instruction in the source language): ``Translate the following text from English to Lithuanian.
Output only the translation, nothing else.'' and ``I\v{s}verskite \v{s}\k{i} tekst\k{a} i\v{s} lietuvi\k{u} kalbos \k{i}
angl\k{u} kalb\k{a}. Pateikite tik vertim\k{a}, nieko daugiau.'' Generation is greedy with at most 256 new tokens.

\textbf{Thinking off.} Qwen3.6 receives an empty \texttt{<think></think>} block, GLM-4.7-Flash's prompt ends in
\texttt{</think>}, Gemma~4's template always adds an empty thought channel, and Qwen3-30B-A3B-Instruct-2507 has no
thinking mode.

\textbf{Gemma~4 patches.} Gemma~4's router has a projection (\texttt{router.proj}) and a per-expert scale
(\texttt{per\_expert\_scale}); both are pruned together with the experts, and \texttt{top\_k\_experts}, Gemma~4's name for the number
of experts per token, is added to the configuration keys that llm-compressor treats as top-$k$. llm-compressor unties Gemma~4's \texttt{lm\_head} from the embeddings, so its pruned
checkpoints store that matrix twice; the duplicate is subtracted from the parameter counts, and loading re-ties the
weights, so evaluation is unaffected.

\textbf{Sequential pipeline and saliency-only runs.} llm-compressor calibrates layer by layer on the already pruned
model, so each configuration is pruned from the original checkpoint; the 24 saliency-only runs (ratio 0) score every
layer on the unpruned model and give the EN/LT saliency correlations. The removed sets of the pruning runs agree with
the lowest-saliency experts of the saliency-only runs with Jaccard \V{seq-25-min}--\V{seq-25-max} at 25\% and
\V{seq-50-min}--\V{seq-50-max} at 50\%.

\textbf{Saliency and selection counts.} Because REAP averages over routed tokens, saliency correlates only
moderately with how often an expert is selected (median within-layer Spearman $\rho=\V{cs-median}$ in the
saliency-only runs). Experts that no calibration token selects get saliency 0 and are removed first; this happens in
\V{never-n} of the 48 configurations, for up to \V{never-max}\% of the removed experts (by model in
\appref{sec:extra}).

\textbf{Uniform pruning.} Every MoE layer loses the same share of its routed experts (25\% or 50\%), so the number of
removed experts per layer is constant by construction; where EN and LT calibration diverge by layer is shown in
Figure~\ref{fig:sallayers} and Table~\ref{tab:depth}.

\textbf{Metrics.} chrF++ is sacrebleu 2.6.0 \texttt{CHRF(word\_order=2)}; BLEU uses the 13a tokeniser. Bits per byte is
NLL per UTF-8 byte of the scored text divided by $\ln 2$. Router entropy uses the softmax over router logits for Qwen
and Gemma but GLM's sigmoid scores normalised to sum to 1, so GLM's router entropy is not strictly comparable. The
off-target rate counts an output as being in the wrong language when it contains more words that occur only in the
other language's FLORES references and unpruned outputs than words that occur only in its own. The unseen-word split
classifies words of terminated EN$\to$LT outputs (after collapsing immediate repeats of 1--4-word sequences) that occur
in no Lithuanian reference or unpruned Lithuanian output: English or source words (in the English vocabulary or the
source sentence), words with letters outside the Lithuanian alphabet, and the remaining Lithuanian-looking forms.
Leaked Gemma~4 reasoning is detected as an output starting with ``thought'', ``Thinking Process'', \texttt{<think>} or a
channel tag; the detector flags no unpruned output. \emph{Selection entropy} is the entropy of a layer's
expert-selection frequencies divided by its maximum, and EN--LT routing divergence is the Jensen--Shannon divergence
(JSD; \citealp{lin1991divergence}) of the two languages' selection frequencies. The Jaccard index of two expert sets is
the size of their intersection divided by the size of their union; random selection gives \V{jac-random-25} at 25\%
and \V{jac-random-50} at 50\% pruning for the kept sets, and \V{remjac-random-25} at 25\% for the removed sets. Per-token
perplexity is not comparable across languages because Lithuanian is split into more tokens (unpruned Qwen3-30B:
\V{ppl-qthree-lt} on Lithuanian, \V{ppl-qthree-en} on English Wikipedia). Statistics use percentile bootstrap intervals
with seed 0 and 10{,}000 resamples (2{,}000 for FLORES gaps and paired effects); McNemar tests \citep{mcnemar1947note} compare each pruned model's answers with those of its unpruned model,
and their $p$-values were also adjusted with Benjamini--Hochberg: \V{a2-mcnemar-raw}/\V{a2-n} run$\times$language pairs
are significant before and \V{a2-mcnemar-bh} after adjustment.

\textbf{Software and hardware.} Pruning used llm-compressor 0.15.0, transformers 5.21.0 and torch 2.14.1; evaluation
used transformers 5.21.1, torch 2.14.1 (CUDA 13.0) and sacrebleu 2.6.0, on 4 NVIDIA RTX PRO 6000 Blackwell Server
Edition GPUs (96\,GB each). Pruning one configuration took \V{prunetime-min-min}--\V{prunetime-min-max} minutes. The
per-item outputs and analysis scripts are available on request (see Data and code availability).

\textbf{Compression.} Pruning saves \V{gb-saved-25-min}--\V{gb-saved-25-max}\,GB per model at 25\%
(Table~\ref{tab:compression}).

\begin{table}[htbp]
\centering
\caption{Compression per model and ratio (means over the six calibration sets; prune time is the range). Parameter
counts include the vision towers of Gemma~4 and Qwen3.6 and exclude the MTP heads; checkpoint sizes include the MTP
heads, which are copied unpruned.}
\label{tab:compression}
\footnotesize
\setlength{\tabcolsep}{3pt}
\resizebox{\textwidth}{!}{\begin{tabular}{@{}llrrrrrr@{}}
\toprule
Model & Ratio & experts/layer & params (B) & params $-$\% & size (GB) & size $-$\% & prune time (min) \\
\midrule
Qwen3-30B-A3B-Instruct-2507 & 25\% & 128$\to$96 & 30.5$\to$23.3 & 23.7 & 61.1$\to$46.6 & 23.8 & 29--30 \\
 & 50\% & 128$\to$64 & 30.5$\to$16.0 & 47.5 & 61.1$\to$32.1 & 47.5 & 27 \\
Qwen3.6-35B-A3B & 25\% & 256$\to$192 & 34.7$\to$26.6 & 23.2 & 71.9$\to$55.8 & 22.4 & 51--52 \\
 & 50\% & 256$\to$128 & 34.7$\to$18.5 & 46.5 & 71.9$\to$39.7 & 44.8 & 46--47 \\
gemma-4-26B-A4B-it & 25\% & 128$\to$96 & 25.8$\to$20.1 & 22.1 & 51.6$\to$40.2 & 22.1 & 21 \\
 & 50\% & 128$\to$64 & 25.8$\to$14.4 & 44.3 & 51.6$\to$28.8 & 44.3 & 20 \\
GLM-4.7-Flash & 25\% & 64$\to$48 & 29.9$\to$23.0 & 23.2 & 62.4$\to$48.5 & 22.3 & 17 \\
 & 50\% & 64$\to$32 & 29.9$\to$16.0 & 46.4 & 62.4$\to$34.7 & 44.5 & 15 \\
\bottomrule
\end{tabular}
}
\end{table}

\FloatBarrier
\section{LT--EN gaps in detail}\label{sec:full}

\begin{figure}[htbp]
\includegraphics[width=\textwidth]{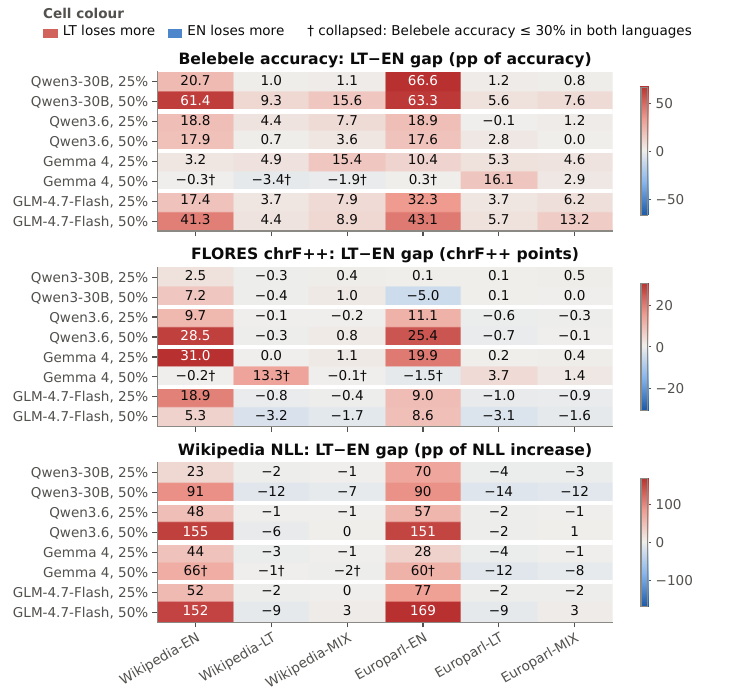}
\caption{\textbf{LT--EN gap for all 48 configurations.} Panels from top: Belebele (pp), FLORES (chrF++), Wikipedia
(pp); rows: model and ratio; columns: calibration set. Red =
Lithuanian loses more, blue = English loses more; Wikipedia values rounded to integers. $\dagger$ = collapsed in both
languages (Belebele accuracy $\le$30\%): a gap near 0 there means that both languages broke, not that they degraded
equally. The EN-calibrated columns are red throughout.}
\label{fig:gapheatmap}
\end{figure}

Figure~\ref{fig:gapheatmap} gives the LT--EN gap of every configuration.

\subsection*{Details for Section~\ref{sec:rq1}}
\begin{itemize}
\item \emph{Significance.} Over both ratios, the EN-calibrated gap is significantly positive in \V{sigpos-bel-en}/16
Belebele, \V{sigpos-flo-en}/16 FLORES and \V{sigpos-wiki-en}/16 Wikipedia configurations, and the LT and MIX Belebele
gaps in \V{sigpos-bel-lt}/16 and \V{sigpos-bel-mix}/16. Under LT and MIX calibration the FLORES gap is significantly
positive in \V{sigpos-flo-lt}/16 and \V{sigpos-flo-mix}/16 configurations and significantly negative in
\V{signeg-flo-lt}/16 and \V{signeg-flo-mix}/16; the Wikipedia gap is significantly negative in \V{signeg-wiki-lt}/16
and \V{signeg-wiki-mix}/16.
\item \emph{Examples.} The Lithuanian accuracy of Qwen3-30B with \emph{Europarl-EN, 25\%} (\V{q3euen25-lt}\%) does not
differ significantly from chance (binomial $p=\V{binom-p}$). Gemma~4 with \emph{Wikipedia-EN, 25\%} also loses
\V{gwien25-en-loss}\pp{} in English, so its gap is only \V{gwien25-gap}\pp.
\item \emph{Strict scoring.} EN-calibrated models at 25\% have a mean argmax accuracy of \V{acc-en-25-en}\% (EN) and
\V{acc-en-25-lt}\% (LT) and a strict accuracy of \V{strict-en-25-en}\% and \V{strict-en-25-lt}\%; the unpruned models
have a strict accuracy of \V{strict-base-en}\% and \V{strict-base-lt}\%.
\item \emph{Ceiling check.} On items that both languages answer correctly when unpruned, LT-calibrated models at 25\%
lose \V{item-lt-25-lt}\% of the Lithuanian and \V{item-lt-25-en}\% of the English answers, more for Lithuanian in
\V{item-lt-25-ltmore}/8 configurations.
\item \emph{Remaining gap by model} (mean Belebele gap under LT and MIX calibration): Qwen3.6 \V{resid-qsix-lt} and
\V{resid-qsix-mix}\pp, the other models \V{resid-qthree-lt}--\V{resid-glm-mix}\pp.
\item \emph{Collapse.} Gemma~4 at 50\% reaches only \V{g50-lt-min}--\V{g50-lt-max}\% in Lithuanian with every
calibration set. The only significantly negative Belebele gap is Gemma~4 with \emph{Wikipedia-LT, 50\%}; the two
significantly negative EN-calibrated FLORES gaps are Qwen3-30B and Gemma~4 with \emph{Europarl-EN, 50\%}.
\end{itemize}

\FloatBarrier
\section{Calibration effects in detail}\label{sec:effects}

\begin{figure}[htbp]
\includegraphics[width=\textwidth]{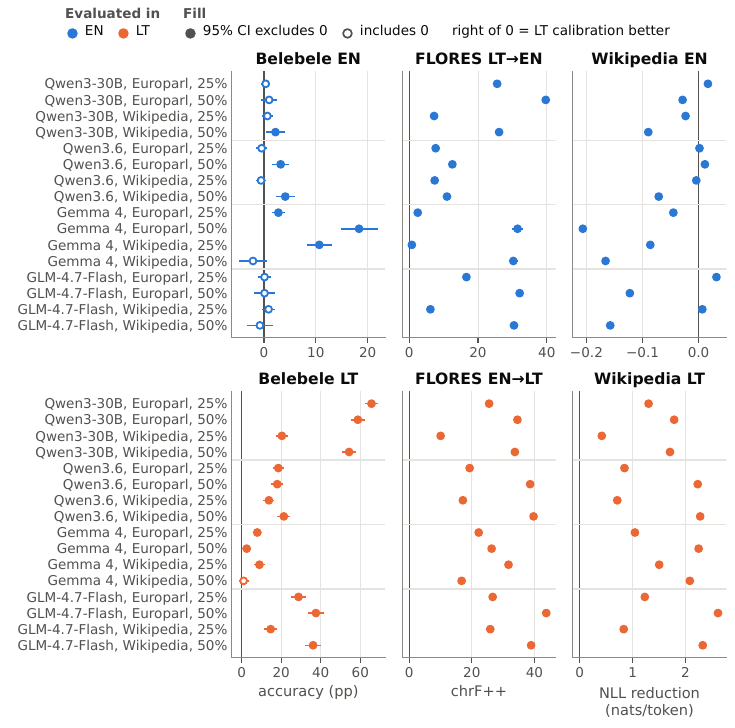}
\caption{\textbf{LT vs EN calibration}, paired by model, domain and ratio (rows). Columns: benchmarks; top row
evaluated in English, bottom row in Lithuanian. Right of 0 = LT calibration better
(Wikipedia NLL negated, nats/token); bars are 95\% intervals; here fill = significance (filled = interval excludes 0),
not ratio; colour = evaluated language. LT calibration is better for every Lithuanian evaluation except collapsed
Gemma~4 \emph{Wikipedia, 50\%}; English Belebele never gets significantly worse.}
\label{fig:forestlten}
\end{figure}

Figure~\ref{fig:forestlten} shows the effect of LT versus EN calibration for every configuration. Table~\ref{tab:eta}
gives the share of variance each design factor explains,
Table~\ref{tab:selection} the calibration set that English- or Lithuanian-only evaluation would select, and
Table~\ref{tab:h3} the change of the gap from 25\% to 50\% pruning.

\subsection*{Details for Section~\ref{sec:rq2}}
\begin{itemize}
\item \emph{Variance shares} (\tabref{tab:eta}). For Lithuanian Belebele degradation, the calibration language explains
\V{eta-ltdeg-lang}\%, the model \V{eta-ltdeg-model}\% and the ratio \V{eta-ltdeg-ratio}\%.
\item \emph{LT versus EN calibration, Lithuanian evaluations.} Belebele is better in \V{eff-lten-bel-lt-better}/16
configurations and worse in \V{eff-lten-bel-lt-worse} (mean gain \V{eff25-lten-bel-lt}\pp{} at 25\%); FLORES is better
in \V{eff-lten-enlt-better}/16 (EN$\to$LT) and \V{eff-lten-lten-better}/16 (LT$\to$EN); Lithuanian NLL is better in
\V{eff-lten-wiki-lt-better}/16.
\item \emph{By model} (mean Lithuanian Belebele gain over the model's four configurations): Qwen3-30B
\V{ml-bel-lt-qthree}, GLM-4.7-Flash \V{ml-bel-lt-glm}, Qwen3.6 \V{ml-bel-lt-qsix} and Gemma~4 \V{ml-bel-lt-gemma}\pp{}
(Gemma~4 collapses at 50\% whatever the calibration); EN$\to$LT gains \V{ml-enlt-gemma}--\V{ml-enlt-glm} chrF++ in
every model.
\item \emph{English under LT calibration.} The median English Belebele effect is \V{ni-median-u}\pp, and the lowest
interval bound over the 16 configurations is \V{ni-worst-lo}\pp. English NLL is worse in
\V{eff-lten-wiki-en-worse}/16 configurations and better in \V{eff-lten-wiki-en-better}/16.
\item \emph{Transfer at 50\%.} The relative Lithuanian NLL increase is \V{relnll-lt-50-lt}\% under LT and
\V{relnll-en-50-lt}\% under EN calibration; the English one is \V{relnll-en-50-en}\% under EN and
\V{relnll-lt-50-en}\% under LT. Qwen3.6 keeps EN$\to$LT at \V{q6lt50-enlt-min}--\V{q6lt50-enlt-max} chrF++ under LT
calibration even at 50\% (unpruned \V{q6-enlt-base}).
\item \emph{MIX versus LT} (positive = MIX better). Lithuanian Belebele \V{eff-mixlt-bel-lt}\pp{} (worse in
\V{eff-mixlt-bel-lt-worse}/16), EN$\to$LT \V{eff-mixlt-enlt} chrF++ (worse in \V{eff-mixlt-enlt-worse}/16), Lithuanian
NLL worse in \V{eff-mixlt-wiki-lt-worse}/16 and English NLL better in \V{eff-mixlt-wiki-en-better}/16 configurations.
MIX gives the smaller absolute Wikipedia gap in \V{h5-absgap-wiki-mix-smaller}/16 pairs.
\item \emph{Choosing by English.} After removing each model's mean, English and Lithuanian Belebele rank the
calibration sets only weakly alike (Kendall $\tau=\V{tau-centred-25}$ at 25\%, $p=\V{tau-centred-25-p}$).
\item \emph{Best compromise.} Qwen3.6 at 25\% has a Belebele degradation (EN/LT) of
\V{best-lt-bel-en}/\V{best-lt-bel-lt}\pp{} with \emph{Europarl-LT} and \V{best-mix-bel-en}/\V{best-mix-bel-lt}\pp{} with
\emph{Europarl-MIX} (negative = better than unpruned).
\end{itemize}

\subsection*{Details for Section~\ref{sec:rq3}}
\begin{itemize}
\item \emph{50\% versus 25\%}. 50\% is significantly worse in
\V{ratio-enlt-worse}/\V{ratio-enlt-n} (EN$\to$LT), \V{ratio-lten-worse}/\V{ratio-lten-n} (LT$\to$EN),
\V{ratio-wiki-en-worse}/\V{ratio-wiki-en-n} and \V{ratio-wiki-lt-worse}/\V{ratio-wiki-lt-n} (Wikipedia EN, LT), and
\V{ratio-bel-en-worse}/\V{ratio-bel-en-n} and \V{ratio-bel-lt-worse}/\V{ratio-bel-lt-n} (Belebele EN, LT)
configurations.
\item \emph{Change of the gap} (Table~\ref{tab:h3}). Under EN calibration, the Belebele gap grows significantly in
\V{h3-belebele-en-up}/8 configurations and shrinks in \V{h3-belebele-en-down}/8, and the FLORES gap grows in
\V{h3-flores-en-up}/8 and shrinks in \V{h3-flores-en-down}/8. Under LT calibration, the Belebele gap grows in
\V{h3-belebele-lt-up}/8 and the Wikipedia gap shrinks in \V{h3-wikipedia-lt-down}/8.
\end{itemize}

\subsection*{Details for Section~\ref{sec:rq4}}
\begin{itemize}
\item Wikipedia calibration lowers Wikipedia NLL in \V{ds-wiki-en-wikipedia}/\V{ds-wiki-en-n} English and
\V{ds-wiki-lt-wikipedia}/\V{ds-wiki-lt-n} Lithuanian configurations, by \V{ds-wiki-signed}\,nats on average, and helps
FLORES LT$\to$EN by \V{ds-lten-mean-u} chrF++ (significant in \V{ds-lten-wikipedia}/24; the reverse in
\V{ds-lten-europarl}/24). Europarl calibration helps English Belebele significantly in \V{ds-bel-en-europarl}/24
configurations (the reverse in \V{ds-bel-en-wikipedia}/24).
\item Under EN calibration at 25\%, GLM-4.7-Flash reaches \V{glmeuen25-lt}\% Lithuanian Belebele accuracy with
\emph{Europarl-EN} and \V{glmwien25-lt}\% with \emph{Wikipedia-EN}.
\item Under LT and MIX calibration the domain still changes Lithuanian Belebele significantly in
\V{ds-bel-lt-lt-sig}/8 and \V{ds-bel-lt-mix-sig}/8 configurations, by up to \V{ds-bel-lt-lt-maxabs} and
\V{ds-bel-lt-mix-maxabs}\pp{} (mostly Gemma~4 and Qwen3.6).
\end{itemize}

\begin{table}[htbp]
\centering
\caption{Share of variance (\%) across the 48 configurations explained by each design factor alone (one-way
$\eta^2$).}
\label{tab:eta}
\footnotesize
\resizebox{\textwidth}{!}{\begin{tabular}{@{}lrrrrr@{}}
\toprule
Factor & LT Belebele degr. & EN Belebele degr. & Belebele gap & FLORES gap & Wikipedia gap \\
\midrule
calibration language & 24 & 0 & 39 & 36 & 70 \\
model & 22 & 50 & 15 & 8 & 4 \\
pruning ratio & 21 & 19 & 1 & 0 & 4 \\
calibration domain & 0 & 1 & 1 & 1 & 0 \\
\bottomrule
\end{tabular}
}
\end{table}

\begin{table}[htbp]
\centering
\caption{Calibration set that maximises English or Lithuanian Belebele accuracy per model and ratio, the Lithuanian
accuracy lost by choosing with English, and the English accuracy given up by choosing with Lithuanian.}
\label{tab:selection}
\footnotesize
\setlength{\tabcolsep}{3pt}
\resizebox{\textwidth}{!}{\begin{tabular}{@{}llllrr@{}}
\toprule
Model & Ratio & best by EN Belebele & best by LT Belebele & LT acc.\ lost (pp) & EN acc.\ cost (pp) \\
\midrule
Qwen3-30B-A3B-Instruct-2507 & 25\% & Europarl-LT & Europarl-MIX & 0.1 & 0.3 \\
 & 50\% & Europarl-MIX & Europarl-LT & 1.6 & 0.4 \\
Qwen3.6-35B-A3B & 25\% & Europarl-MIX & Europarl-LT & 0.1 & 1.2 \\
 & 50\% & Wikipedia-LT & Wikipedia-LT & 0.0 & 0.0 \\
gemma-4-26B-A4B-it & 25\% & Europarl-LT & Europarl-LT & 0.0 & 0.0 \\
 & 50\% & Europarl-LT & Europarl-MIX & 0.2 & 13.0 \\
GLM-4.7-Flash & 25\% & Europarl-MIX & Europarl-LT & 0.6 & 2.0 \\
 & 50\% & Europarl-MIX & Europarl-LT & 6.3 & 1.2 \\
\bottomrule
\end{tabular}
}
\end{table}

\begin{table}[htbp]
\centering
\caption{Change of the LT--EN gap from 25\% to 50\% pruning (gap at 50\% minus gap at 25\%) with 95\% paired bootstrap
intervals (Belebele: items resampled jointly in both languages, 4{,}000 resamples; FLORES: sentences, 1{,}000 resamples;
Wikipedia: documents per language, 4{,}000 resamples). Positive = the gap grows.}
\label{tab:h3}
\footnotesize
\setlength{\tabcolsep}{3pt}
\resizebox{\textwidth}{!}{\begin{tabular}{@{}llrrr@{}}
\toprule
Model & Calibration set & Belebele (pp) & FLORES (chrF++) & Wikipedia (pp) \\
\midrule
Qwen3-30B & Wikipedia-EN & \textbf{+40.8} {\scriptsize[36.7, 44.8]} & \textbf{+4.7} {\scriptsize[3.6, 5.8]} & \textbf{+68.0} {\scriptsize[67.3, 68.7]} \\
 & Wikipedia-LT & \textbf{+8.3} {\scriptsize[5.6, 11.3]} & $-$0.1 {\scriptsize[$-$0.9, 0.6]} & \textbf{$-$9.6} {\scriptsize[$-$9.9, $-$9.4]} \\
 & Wikipedia-MIX & \textbf{+14.4} {\scriptsize[11.3, 17.7]} & +0.6 {\scriptsize[$-$0.1, 1.4]} & \textbf{$-$5.7} {\scriptsize[$-$6.0, $-$5.4]} \\
 & Europarl-EN & \textbf{$-$3.2} {\scriptsize[$-$6.1, $-$0.3]} & \textbf{$-$5.1} {\scriptsize[$-$5.9, $-$4.1]} & \textbf{+19.6} {\scriptsize[19.0, 20.2]} \\
 & Europarl-LT & \textbf{+4.3} {\scriptsize[1.9, 6.9]} & 0.0 {\scriptsize[$-$0.7, 0.7]} & \textbf{$-$10.5} {\scriptsize[$-$10.9, $-$10.2]} \\
 & Europarl-MIX & \textbf{+6.8} {\scriptsize[4.2, 9.3]} & $-$0.5 {\scriptsize[$-$1.2, 0.3]} & \textbf{$-$9.0} {\scriptsize[$-$9.4, $-$8.6]} \\
\midrule
Qwen3.6 & Wikipedia-EN & $-$0.9 {\scriptsize[$-$4.8, 2.9]} & \textbf{+18.8} {\scriptsize[17.5, 20.1]} & \textbf{+107.3} {\scriptsize[106.3, 108.4]} \\
 & Wikipedia-LT & \textbf{$-$3.8} {\scriptsize[$-$6.2, $-$1.2]} & $-$0.2 {\scriptsize[$-$0.8, 0.3]} & \textbf{$-$4.6} {\scriptsize[$-$4.9, $-$4.4]} \\
 & Wikipedia-MIX & \textbf{$-$4.1} {\scriptsize[$-$6.9, $-$1.3]} & \textbf{+1.0} {\scriptsize[0.4, 1.5]} & \textbf{+0.6} {\scriptsize[0.3, 0.9]} \\
 & Europarl-EN & $-$1.3 {\scriptsize[$-$5.0, 2.6]} & \textbf{+14.3} {\scriptsize[13.1, 15.5]} & \textbf{+94.6} {\scriptsize[93.6, 95.6]} \\
 & Europarl-LT & +2.9 {\scriptsize[$-$0.1, 5.9]} & $-$0.2 {\scriptsize[$-$0.8, 0.4]} & \textbf{$-$0.6} {\scriptsize[$-$1.0, $-$0.1]} \\
 & Europarl-MIX & $-$1.2 {\scriptsize[$-$4.4, 1.9]} & +0.1 {\scriptsize[$-$0.4, 0.7]} & \textbf{+1.5} {\scriptsize[1.0, 2.1]} \\
\midrule
Gemma 4 & Wikipedia-EN & $-$3.6 {\scriptsize[$-$9.7, 2.4]} & \textbf{$-$31.1} {\scriptsize[$-$32.2, $-$30.1]} & \textbf{+21.6} {\scriptsize[21.2, 22.0]} \\
 & Wikipedia-LT & \textbf{$-$8.3} {\scriptsize[$-$13.4, $-$3.0]} & \textbf{+13.3} {\scriptsize[11.8, 14.9]} & \textbf{+2.0} {\scriptsize[1.7, 2.2]} \\
 & Wikipedia-MIX & \textbf{$-$17.3} {\scriptsize[$-$23.1, $-$11.3]} & \textbf{$-$1.2} {\scriptsize[$-$2.5, $-$0.1]} & \textbf{$-$0.6} {\scriptsize[$-$0.8, $-$0.3]} \\
 & Europarl-EN & \textbf{$-$10.1} {\scriptsize[$-$14.9, $-$5.2]} & \textbf{$-$21.4} {\scriptsize[$-$22.7, $-$20.2]} & \textbf{+32.1} {\scriptsize[31.5, 32.6]} \\
 & Europarl-LT & \textbf{+10.8} {\scriptsize[6.1, 15.7]} & \textbf{+3.4} {\scriptsize[1.8, 5.1]} & \textbf{$-$8.6} {\scriptsize[$-$8.8, $-$8.3]} \\
 & Europarl-MIX & $-$1.7 {\scriptsize[$-$6.7, 3.2]} & +1.0 {\scriptsize[$-$0.7, 2.6]} & \textbf{$-$6.2} {\scriptsize[$-$6.4, $-$5.9]} \\
\midrule
GLM-4.7-Flash & Wikipedia-EN & \textbf{+23.9} {\scriptsize[18.7, 28.9]} & \textbf{$-$13.6} {\scriptsize[$-$14.8, $-$12.4]} & \textbf{+99.3} {\scriptsize[98.4, 100.3]} \\
 & Wikipedia-LT & +0.8 {\scriptsize[$-$3.2, 4.7]} & \textbf{$-$2.3} {\scriptsize[$-$3.1, $-$1.6]} & \textbf{$-$6.7} {\scriptsize[$-$7.2, $-$6.2]} \\
 & Wikipedia-MIX & +1.0 {\scriptsize[$-$2.7, 4.4]} & \textbf{$-$1.3} {\scriptsize[$-$2.2, $-$0.6]} & \textbf{+3.2} {\scriptsize[2.6, 3.8]} \\
 & Europarl-EN & \textbf{+10.8} {\scriptsize[6.6, 15.0]} & $-$0.4 {\scriptsize[$-$1.6, 0.6]} & \textbf{+91.6} {\scriptsize[90.6, 92.6]} \\
 & Europarl-LT & +2.0 {\scriptsize[$-$1.4, 5.6]} & \textbf{$-$2.1} {\scriptsize[$-$2.8, $-$1.4]} & \textbf{$-$6.7} {\scriptsize[$-$7.4, $-$6.1]} \\
 & Europarl-MIX & \textbf{+7.0} {\scriptsize[3.4, 10.4]} & $-$0.8 {\scriptsize[$-$1.6, 0.1]} & \textbf{+4.8} {\scriptsize[4.2, 5.5]} \\
\bottomrule
\end{tabular}
}
\end{table}

\FloatBarrier
\section{Experts and routing}\label{sec:extra}

\begin{figure}[htbp]
\includegraphics[width=\textwidth]{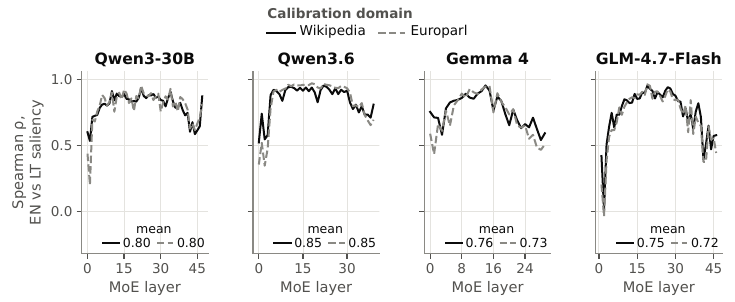}
\caption{\textbf{Per-layer EN vs LT saliency rank correlation} (saliency-only runs; solid = Wikipedia, dashed =
Europarl). Agreement is lowest in the first and last layers.}
\label{fig:sallayers}
\end{figure}

\begin{figure}[htbp]
\includegraphics[width=\textwidth]{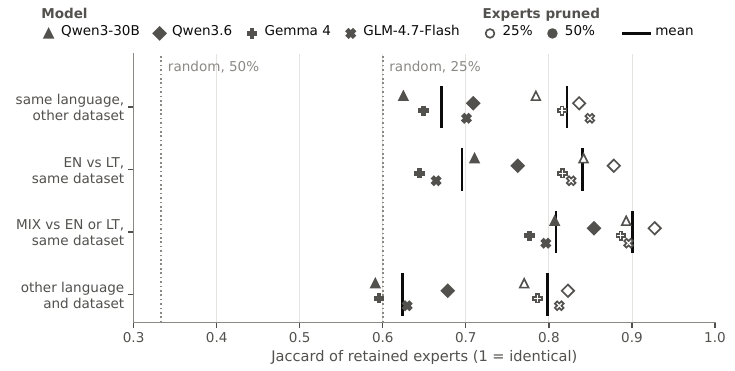}
\caption{\textbf{Which calibration changes move the retained expert set?} Jaccard of retained experts per kind of
calibration pair, averaged per model (glyph = model); hollow = 25\%, filled = 50\%; black bars are means; dotted lines are random
selection. Changing the domain moves the set at least as much as changing EN to LT in three of the four models.}
\label{fig:overlappairs}
\end{figure}

\textbf{Saliency and retention.} EN and LT calibration rank experts similarly but not identically, with the lowest
agreement in the first and last layers (Figure~\ref{fig:sallayers}); Figure~\ref{fig:overlappairs} shows which calibration
changes move the retained expert set, and Table~\ref{tab:depth} shows that Lithuanian text is routed more narrowly
than English text.

\subsection*{Details for Section~\ref{sec:rq5}}
\begin{figure}[htbp]
\includegraphics[width=\textwidth]{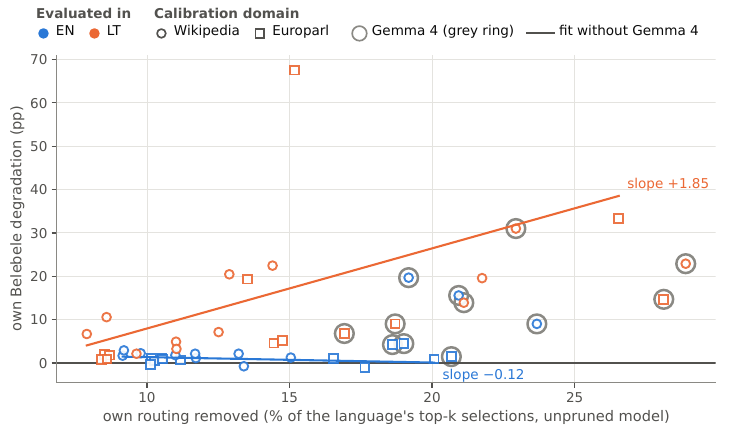}
\caption{\textbf{Coverage vs robustness.} Each language's own routing share removed (x) against its own Belebele
degradation (y; positive = worse), every configuration at 25\%; colour = evaluated language; circle/square =
Wikipedia/Europarl; grey rings mark Gemma~4, which is excluded from the fitted lines. Lithuanian accuracy falls
steeply with the share of its routing removed, English accuracy barely moves.}
\label{fig:coverage}
\end{figure}

\begin{itemize}
\item \emph{Routing of the unpruned models} (ranges over the four models). Normalised selection entropy is
\V{selent-lt-min}--\V{selent-lt-max} for Lithuanian and \V{selent-en-min}--\V{selent-en-max} for English text, and the
EN--LT JSD is \V{jsd-min}--\V{jsd-max}.
\item \emph{Overlap of expert sets} (same domain, EN versus LT calibration). The Jaccard index of the kept sets is
\V{jac-enlt-25-min}--\V{jac-enlt-25-max} at 25\% (random \V{jac-random-25}) and \V{jac-enlt-50-min}--\V{jac-enlt-50-max}
at 50\% (random \V{jac-random-50}); that of the removed sets is \V{remjac-enlt-25-min}--\V{remjac-enlt-25-max} at 25\%
(random \V{remjac-random-25}).
\item \emph{Removed routing.} Under EN calibration the removed experts had received \V{remmass-ratio}$\times$ as large a
share of Lithuanian as of English routing at 25\%, and \V{remmass-en-50-lt}\% against \V{remmass-en-50-en}\%
(\V{remmass-ratio-50}$\times$) at 50\%. In Figure~\ref{fig:coverage}, the English slope without Gemma~4 is
\V{cov-slope-en}.
\item \emph{Depth} (Table~\ref{tab:depth}). The EN--LT JSD is \V{depth-middle-jsd-min}--\V{depth-middle-jsd-max} in the
middle layers, where expert usage ranks agree ($\rho=\V{depth-middle-rho-min}$--\V{depth-middle-rho-max}), and
\V{depth-first-jsd-min}--\V{depth-first-jsd-max} and \V{depth-last-jsd-min}--\V{depth-last-jsd-max} in the first and
last 15\% of MoE layers. In the last layers, LT calibration removes \V{top8-last-ltcal-lt-min}--\V{top8-last-ltcal-lt-max}
of Lithuanian's 8 most-used experts per layer. The relative LT--EN difference in removed routing under EN calibration
is largest there (\V{bandmass-en-last-lt}\% against \V{bandmass-en-last-en}\%), mainly because the English share
removed is lowest.
\item \emph{Depth by model.} Under EN calibration, the extra Lithuanian share removed peaks in the last 15\% of layers
for GLM-4.7-Flash (\V{extra-glm-last}\pp, against \V{extra-glm-middle} in the middle layers) and Qwen3.6
(\V{extra-qsix-last} against \V{extra-qsix-middle}), at 65--85\% of depth for Gemma~4 (\V{extra-gemma-late}\pp), and is
small and even for Qwen3-30B (\V{extra-qthree-min}--\V{extra-qthree-max}\pp), the model with the worst collapse.
\item \emph{Correlations with Lithuanian Belebele loss} at 25\% ($n=24$). For the Lithuanian routing removed in all
layers $\rho=\V{r12-all_25-lt-ltdeg}$, in the first 15\% $\rho=\V{r12-all_25-lt_first-ltdeg}$ and in the last 15\%
$\rho=\V{r12-all_25-lt_last-ltdeg}$. Within EN calibration, the extra Lithuanian share removed does not track the
Belebele gap ($\rho=\V{r12-en_all-extra_lt-bel}$, $n=16$).
\item \emph{Overlap and gaps}. The correlation of EN--LT overlap with the Belebele gap
is $\rho=\V{ovl-bel-pooled-rho}$ ($n=16$). The two pooled correlations with $p<0.05$ (BLEU gap,
$\rho=\V{ovl-bleu-pooled-rho}$; Wikipedia gap, $\rho=\V{ovl-wiki-pooled-rho}$) fail a Bonferroni correction for the 10
tests and vanish within each ratio ($|\rho|\le\V{ovl-25-maxabs}$ at 25\%).
\end{itemize}

\begin{table}[htbp]
\caption{\textbf{Where in the network English and Lithuanian differ.} Left: routing of the unpruned models, as ranges
over the four models: EN$-$LT difference in normalised selection entropy ($\Delta H$), routing divergence (JSD), rank
correlation of expert usage ($\rho$), and share of routing handled by the 8 most-used experts (\%). Right: share of each
language's routing on the experts removed at 25\% (\%, mean of 8 configurations). Depth bands: first 15\%, middle
35--65\% and last 15\% of the MoE layers; the measures are defined in \appref{sec:impl}.}
\label{tab:depth}
\footnotesize
\setlength{\tabcolsep}{3pt}
\resizebox{\textwidth}{!}{\begin{tabular}{@{}lrrrrrrrrr@{}}
\toprule
 & \multicolumn{5}{c}{Unpruned routing, EN vs LT (range over 4 models)} & \multicolumn{4}{c}{On removed experts at 25\% (\%)} \\
\cmidrule(lr){2-6}\cmidrule(lr){7-10}
 & & & & \multicolumn{2}{c}{top-8 share} & \multicolumn{2}{c}{EN calib.} & \multicolumn{2}{c}{LT calib.} \\
\cmidrule(lr){5-6}\cmidrule(lr){7-8}\cmidrule(lr){9-10}
Depth & $\Delta H$ & JSD & $\rho$ & EN & LT & EN & LT & EN & LT \\
\midrule
first 15\% & 0.07--0.12 & 0.16--0.28 & 0.45--0.73 & 14--32 & 28--49 & 20.4 & 25.1 & 22.8 & 16.1 \\
middle 35--65\% & 0.03--0.07 & 0.09--0.13 & 0.84--0.91 & 22--41 & 29--53 & 13.7 & 18.7 & 14.6 & 14.0 \\
last 15\% & 0.14--0.22 & 0.33--0.49 & 0.21--0.69 & 19--39 & 55--72 & 10.3 & 20.5 & 14.8 & 8.1 \\
\bottomrule
\end{tabular}
}
\end{table}

\textbf{Saliency concentration.} Lithuanian-calibrated saliency is more concentrated (higher Gini) than
English-calibrated saliency in \V{a6-ltgini-higher-pct}\% of layers (Qwen3-30B \V{a6-ltgini-higher-qthree}\%, Gemma~4
\V{a6-ltgini-higher-gemma}\%, GLM-4.7-Flash \V{a6-ltgini-higher-glm}\%, Qwen3.6 \V{a6-ltgini-higher-qsix}\%). \textbf{Never-routed experts} (saliency 0, removed first) are more common under LT calibration:
\V{never-lt-25}/8 configurations at 25\% vs \V{never-en-25}/8 under EN; by model, Gemma~4 \V{never-gemma-n}/12 (up to
\V{never-gemma-max}\% of removed experts), Qwen3-30B \V{never-qthree-n}/12 (\V{never-qthree-max}\%), Qwen3.6
\V{never-qsix-n}/12 (\V{never-qsix-max}\%), GLM-4.7-Flash \V{never-glm-n}/12 (\V{never-glm-max}\%). \textbf{Router
entropy after pruning}: normalised selection entropy rises slightly (\V{a7-sel-min} to \V{a7-sel-max}), so routing
spreads over the remaining experts. \textbf{Per-layer links}: within each model, layers with
lower EN--LT saliency correlation retain less similar expert sets ($\rho$ = \V{a1-sal-min}--\V{a1-sal-max} across
layers; layers are not independent, so we give no $p$-values), and routing divergence correlates positively with the
extra Lithuanian routing removed in all four models (\V{a1-jsdextra-min}--\V{a1-jsdextra-max}).

\FloatBarrier
\section{Failure modes}\label{sec:failures}

Figure~\ref{fig:collapse} shows how Belebele answers collapse onto one letter.

\subsection*{Details for Section~\ref{sec:rq6}}
\begin{figure}[htbp]
\includegraphics[width=\textwidth]{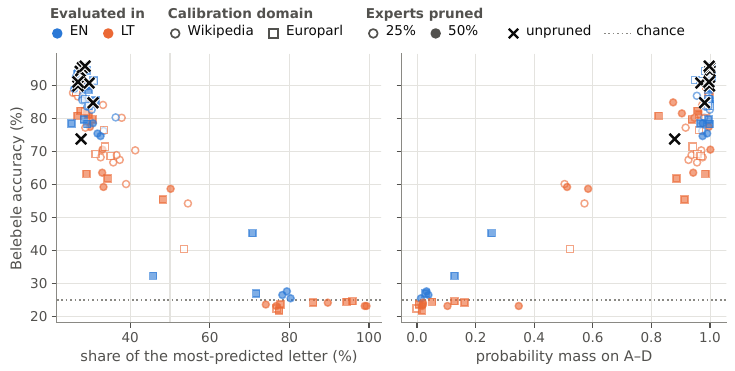}
\caption{\textbf{Belebele accuracy vs answer-letter collapse} for the 48 pruned runs, each evaluated in both
languages (96 points; colour = evaluated language; circle/square = Wikipedia/Europarl; hollow/filled = 25\%/50\%),
and the unpruned models ($\times$). Left: share of the most-predicted letter; right: probability mass on A--D. Runs
at chance accuracy put most answers on one letter and little probability on letters at all.}
\label{fig:collapse}
\end{figure}

\begin{itemize}
\item \emph{Runaway generation.} With \emph{Europarl-EN, 50\%}, GLM-4.7-Flash hits the 256-token limit in
\V{hits-glm-euen50}\% of EN$\to$LT sentences. The limit allows about \V{limit-words-lt-min}--\V{limit-words-lt-max}
Lithuanian but \V{limit-words-en-min}--\V{limit-words-en-max} English words, so hit rates are not comparable across
directions.
\item \emph{Off-target output} (definition in \appref{sec:impl}). Under EN calibration \V{off-en-25-enlt}\% of EN$\to$LT
outputs are off-target at 25\% and \V{off-en-50-enlt}\% at 50\% (up to \V{off-en-50-enlt-max}\% in one configuration);
under MIX calibration \V{off-mix-50-enlt}\% at 50\%; for the unpruned models \V{off-base-max}\%.
\item \emph{Content.} Under EN calibration at 25\%, \V{a4-en-25-enlt-hits}\% of EN$\to$LT outputs hit the limit, and
collapsing immediate repeats of 1--4-word sequences recovers only \V{a4-en-25-enlt-recovered} of the
\V{a4-en-25-enlt-all} chrF++ lost. LT-calibrated translations at 25\% are within \V{a9-ltcal25-enlt-chrfdrop} chrF++ of
the unpruned model, so their \V{split-lt-25-ltlike}\% of unseen Lithuanian-looking words approximates legitimate
variation. English or copied words make up \V{split-en-25-eng}\% of all words under EN calibration at 25\%.
\item \emph{Broken diacritics.} The \V{fffd-total} outputs with invalid byte sequences come from \V{fffd-runs} pruned
runs.
\item \emph{Letter collapse} (Figure~\ref{fig:collapse}). Gemma~4 at 50\% puts at least 70\% of its answers on ``A'' in
\V{g50-A-ge70} of its \V{g50-A-n} run$\times$language pairs, and EN-calibrated GLM-4.7-Flash and Qwen3-30B at 50\% put
\V{encal50-lt-A-min}--\V{encal50-lt-A-max}\% of their Lithuanian answers there. The probability mass on the four
letters falls to \V{g50-mass-min}--\V{g50-mass-max} for Gemma~4 at 50\%. Every collapsed run puts at least
\V{collapsed-top-min}\% of its answers on one letter, and no run above 60\% accuracy more than \V{healthy-top-max}\%.
\item \emph{Reasoning text.} In all \V{thought-50-configs} Gemma~4 configurations at 50\%,
\V{thought-50-lten-min}--\V{thought-50-lten-max} of the 1{,}012 LT$\to$EN outputs open with a reasoning block.
Excluding them changes the LT$\to$EN drop of \emph{Europarl-LT, 50\%} only from \V{a4-g-europarllt50-lten-all} to
\V{a4-g-europarllt50-lten-nothought} chrF++.
\item \emph{Metric checks.} Answer flips track absolute accuracy changes ($\rho=\V{a2-rho}$).
BLEU and chrF++ disagree on the sign of the gap in \V{a3-dis-abs}/48 configurations, only because LT$\to$EN BLEU starts
much higher; relative to each direction's unpruned score they agree ($\rho=\V{a3-rho-rel}$).
\end{itemize}

\FloatBarrier
\end{appendix}

\end{document}